\documentclass[runningheads]{llncs}

\usepackage{eccv}

\usepackage{eccvabbrv}

\usepackage{amsmath}

\DeclareMathOperator*{\argmin}{arg\,min}

\usepackage{graphicx}

\usepackage{tikz}
\usepackage{comment}
\usepackage{color}

\usepackage{enumitem}
\usepackage{graphicx}
\usepackage{amsmath}
\usepackage{booktabs}
\usepackage{mathtools}
\usepackage{url}
\usepackage{comment}
\usepackage{color}
\usepackage{graphicx,setspace}
\usepackage{multirow, array, booktabs,makecell,xspace,bm}
\usepackage{pifont}% http://ctan.org/pkg/pifont
\usepackage{amsfonts}
\usepackage{arydshln}
\usepackage{dsfont}

\usepackage{stackengine}

\usepackage{scalerel}
\usepackage{esdiff} 

\usepackage{bbm}

\newcommand{\cmark}{\ding{51}}%
\usepackage{comment}
\usepackage{color}
\usepackage{graphicx, setspace}
\usepackage{multirow, array, booktabs,makecell,xspace,bm, arydshln}
\usepackage{amsmath,amssymb, pifont}

\usepackage{hyperref}

\usepackage{orcidlink}

\begin{document}

\title{MonoWAD: Weather-Adaptive Diffusion Model for Robust Monocular 3D Object Detection} 

\titlerunning{MonoWAD}

\author{Youngmin Oh\inst{1}\orcidlink{0009-0003-8949-9927} \and
Hyung-Il Kim\inst{2}\orcidlink{0000-0001-6425-549X} \and
Seong Tae Kim\inst{1}\thanks{Corresponding author}\orcidlink{0000-0002-2132-6021} \and
Jung Uk Kim\inst{1\dagger}\orcidlink{0000-0003-4533-4875}}

\authorrunning{Y.~Oh et al.}

\institute{Kyung Hee University, Yong-in, South Korea\\
\email{\{oym9104, st.kim, ju.kim\}@khu.ac.kr}\\
\and ETRI, Daejeon, South Korea\\
\email{hikim@etri.re.kr}}

\maketitle

\begin{abstract}

Monocular 3D object detection is an important challenging task in autonomous driving. Existing methods mainly focus on performing 3D detection in ideal weather conditions, characterized by scenarios with clear and optimal visibility. However, the challenge of autonomous driving requires the ability to handle changes in weather conditions, such as foggy weather, not just clear weather. We introduce MonoWAD, a novel weather-robust monocular 3D object detector with a weather-adaptive diffusion model. It contains two components: (1) the weather codebook to memorize the knowledge of the clear weather and generate a weather-reference feature for any input, and (2) the weather-adaptive diffusion model to enhance the feature representation of the input feature by incorporating a weather-reference feature. This serves an attention role in indicating how much improvement is needed for the input feature according to the weather conditions. To achieve this goal, we introduce a weather-adaptive enhancement loss to enhance the feature representation under both clear and foggy weather conditions. Extensive experiments under various weather conditions demonstrate that MonoWAD achieves weather-robust monocular 3D object detection. The code and dataset are released at \href{https://github.com/VisualAIKHU/MonoWAD}{https://github.com/VisualAIKHU/MonoWAD}.

\end{abstract}

\section{Introduction}
\label{sec:intro}

%------------------------------------ Figure 1 
%###############################################################################%######
\begin{figure}[t]
    \begin{minipage}[b]{1.0\linewidth}
	\centering
        \centerline{\includegraphics[width=12.5cm]{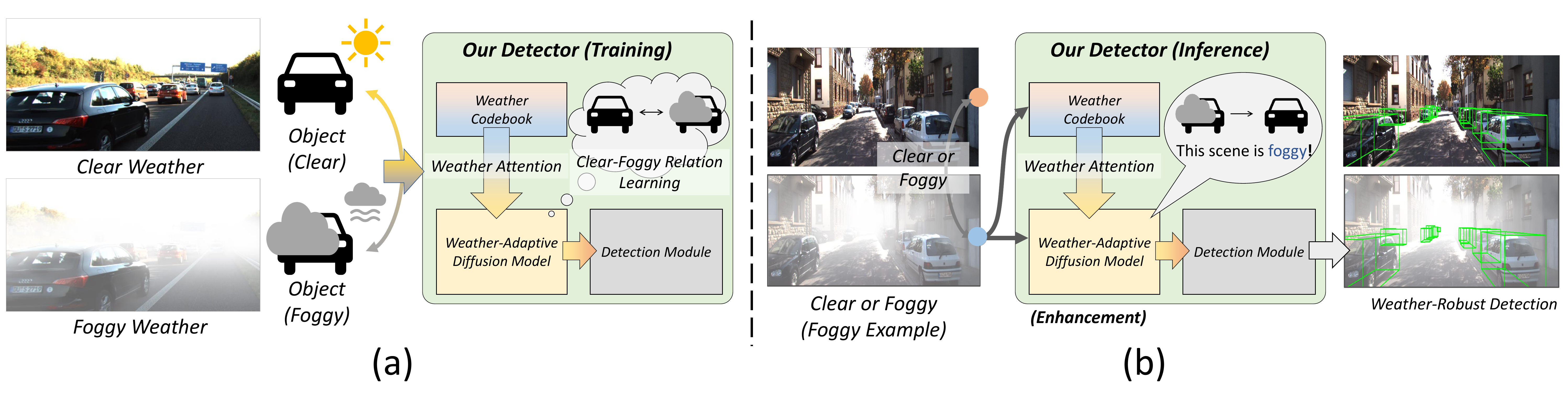}}
        \end{minipage}
        \vspace{-0.8cm}
	\caption{Conceptual diagram of the proposed method (foggy example). (a) In the training phase, weather codebook learns the clear knowledge to transfer it to weather-adaptive diffusion model to enhance content related to the weather conditions. (b) By doing so, even with input images under various weather conditions (\textit{e.g.,} foggy images), monocular 3D object detection becomes adaptable to various weather scenarios.}
    \vspace{-0.4cm}
    \label{fig:1}
\end{figure}
%#####################################################################################

Monocular 3D object detection aims to detect 3D objects using only a single camera \cite{MonoRCNN, memory-3dod, MonoFlex, GUPNet, MonoDTR, MonoDETR}. In contrast to LiDAR-based methods that rely on expensive LiDAR sensors for depth estimation \cite{VoxelNet, PointPillars, PointRCNN, PV-RCNN}, and stereo-based methods that require synchronized stereo cameras, monocular 3D object detection only requires monocular images, offering the advantage of computational cost-effectiveness and requiring fewer resources. Due to this characteristic, the monocular 3D object detection is applied to a wide range of real-world applications, such as autonomous driving \cite{KITTI, Survey_of_Autonomous_Driving,kim2021robust} and robotics \cite{CubeSLAM}.

However, existing monocular 3D object detectors mainly focus on ideal autonomous driving environments (\textit{i.e.,} clear weather). There are challenges in applying them to real-world scenarios with adverse weather conditions, such as fog and rain. Among these, fog poses the most significant challenge compared to other weather \cite{Fog_crash, foggy_data}. This is due to the dense and diffuse nature of fog, which strongly scatters and absorbs light, leading to difficulties in object detection \cite{SeeingThroughFog}. Since monocular 3D object detection relies solely on visual information from a single image, unlike LiDAR, it is crucial to design detectors to achieve enhanced performance in challenging visibility scenarios.

In this paper, we propose MonoWAD, a novel weather-robust monocular 3D object detector to address the aforementioned issues. As mentioned earlier, due to the inherent challenges posed by foggy weather among various adverse weather conditions, we focus primarily on clear and foggy weather (results for other weather conditions such as rainy and sunset are presented in Section~\ref{sec:exper}). For weather-robust object detection, clear weather requires relatively modest improvements to enhance the visual representation of features. In contrast, significant enhancements in this feature representation are required for foggy weather. To address this, we consider two key aspects: (1) how to quantify the degree of improvement needed for the input image, and (2) how to guide the representation of the input image.

To address the two key aspects, as shown in Fig. \ref{fig:1}, our MonoWAD consists of a weather codebook and weather-adaptive diffusion model. First, we introduce the weather codebook to generate a weather-reference feature that contains knowledge about reference weather in a given scene. The reference weather acts as a guide, indicating the degree of weather improvement required. Since the clear weather contains a richer visual representation of objects, we adopt it as a reference weather. At this time, we devise a clear knowledge recalling (CKR) loss to guide the weather codebook to memorize information about clear weather and generate a weather-reference feature for any input (clear or foggy). As a result, our detector can understand where improvements are needed in the input features based on the weather-reference feature.

Second, we propose a weather-adaptive diffusion model to effectively enhance feature representations in accordance with weather conditions. Given input feature (clear or foggy), the weather-adaptive diffusion model dynamically enhances the representation of the input feature based on the weather-reference feature. The weather-reference feature plays a role of attention, determining the extent to which the input features need improvement. At this time, we define the difference between clear and foggy weather (\textit{i.e.,} weather changes) as the fog distribution to adopt it as the noise for our diffusion model. With fog distribution, our weather-adaptive diffusion model can enhance the feature representation according to the weather conditions through multiple steps of reverse processes. To achieve this goal, we introduce a weather-adaptive enhancement (WAE) loss. As a result, our MonoWAD performs weather-robust detection by adaptively improving feature representation according to the weather conditions.

To adaptively enhance the feature representation through the difference between weather conditions, we generate a new foggy KITTI dataset based on the KITTI dataset \cite{KITTI}. Comprehensive experimental results on several datasets \cite{KITTI, Virtual_KITTI1} show that our MonoWAD outperforms the existing state-of-the-art monocular 3D object detectors \cite{GUPNet, DID-M3D, MonoGround, MonoDTR, MonoDETR} under foggy weather, which is the most challenging weather condition. While our method primarily focuses on foggy, experiments conducted under various weather conditions (\textit{e.g.,} foggy, rainy, and sunset) have demonstrated its applicability to other weather scenarios.

The main contributions of our paper can be summarized as follows:
\begin{itemize}
    \item We introduce a new weather-robust monocular 3D object detector, called MonoWAD, that is robust to various weather conditions.
    \item We design a weather codebook with clear knowledge recalling loss for learning about clear weather, providing reference information for enhancement. 
    \item We propose weather-adaptive diffusion model with weather-adaptive enhancement loss to dynamically enhance the feature representation of the input images according to the weather conditions.
\end{itemize}

\section{Related Work}
\subsection{Monocular 3D Object Detection}

Monocular 3D object detection task can be categorized into two directions according to the type of data used in the training phase: (1) using only a monocular image and (2) incorporating additional data, such as depth along with a monocular image. The first category relies on the geometric relationship between 2D and 3D \cite{MonoPair, SMOKE, MonoRCNN, MonoFlex, MonoGround, GUPNet, MonoDETR, MonoCon, Deep3Dbox, M3D-RPN, M3DSSD}. For example, Deep3Dbox~\cite{Deep3Dbox} utilizes the geometric information of 2D bounding boxes to predict 3D bounding boxes. In~\cite{M3D-RPN}, M3D-RPN was proposed to understand the 3D scene from the depth-aware convolution. MonoRCNN~\cite{MonoRCNN} predicted 3D bounding boxes through geometry-based distance decomposition, and MonoCon~\cite{MonoCon} learns mono context for 3D object detection. MonoDETR~\cite{MonoDETR} introduces a depth-guided transformer that utilizes geometric depth cues without requiring additional data.

Moreover, since monocular image contains limited information for estimating 3D object cues, monocular 3D object detectors have adopted the additional data for more robust detection \cite{DDMP, CaDDN, Kinematic3D, D4LCN, DID-M3D, MonoDTR, PatchNet, AutoShape}. For example, the depth-conditioned dynamic message propagation (DDMP) \cite{DDMP} was proposed to integrate prior depth information with the image context. CaDDN~\cite{CaDDN} was introduced to utilize the depth distribution of predicted categories for each pixel to project the context information onto 3D space, deriving 3D bounding boxes. MonoDTR~\cite{MonoDTR} employs the transformer architecture to integrate depth features and context, thus estimating more accurate depth information.
 
Despite recent progress, the existing monocular 3D object detectors mainly rely on the benchmark data collected under clear weather conditions. However, it is essential to account for challenging adverse weather conditions, such as fog, to more accurately reflect real-world scenarios. In this paper, we aim to introduce weather-robust 3D object detection by enhancing visual features with the proposed weather-adaptive diffusion model and weather codebook.
\vspace{-0.3cm}
\subsection{Computer Vision Tasks for Foggy Weather}

There have been a lot of studies on improving performance in various weather conditions for real-world application of computer vision technology~\cite{Foggy_Cityscapes, SeeingThroughFog, Semantic_Segmentation_weather, Fog_3dod, kim2022towards, kim2021uncertainty, Fog_Simulation, Fog_3dod3, DA_foggy}. 
In particular, fog is considered one of the most critical issues due to its significant degradation of visual information compared to other weather conditions. To deal with the foggy weather conditions, the authors in~\cite{Foggy_Cityscapes} generated synthetic fog images (so-called Foggy Cityscapes) from clear weather images, and utilized them for training semantic segmentation and 2D object detection. Similar to Foggy Cityscapes, Martin \textit{et al.}~\cite{Fog_Simulation} naturally synthesizes fog into LIDAR to enhance the performance of LiDAR-based 3D object detectors in foggy weather.  Bijelic \textit{et al.}~\cite{SeeingThroughFog} use real dataset including foggy weather, for 2D detection with the multimodal fusion networks. Mai \textit{et al.}~\cite{Fog_3dod} synthesize fog for LiDAR and stereo images and perform fusion-based 3D object detection using the SLS-Fusion network. Xin \textit{et al.}~\cite{DA_foggy} focused on 2D detection in foggy weather by applying domain adaptation. In this context, we focus on monocular 3D object detection in scenarios that rely solely on visual information in challenging foggy environments. To address this challenge, we propose a weather-robust diffusion model that dynamically improves features based on reference feature.

\vspace{-0.3cm}
\subsection{Diffusion Models}
Recently, the diffusion model~\cite{DDPM} has attracted considerable interest in computer vision due to its impressive progress in image generation~\cite{latent-diffusion, Inpainting, NeuralField_ldm, video_diffusion}, as well as its potential application to other vision tasks such as segmentation~\cite{ddpm_segmentation} and image captioning~\cite{ddpm_captioning}. Inspired by the remarkable generative ability, we design the diffusion model for robust monocular 3D object detection by considering a foggy effect (which is one of the challenging adverse weather conditions for monocular 3D object detection) as a form of noise in the model. That is, we propose a method in which visual features obscured by fog are progressively improved by training a diffusion model based on the forward/reverse diffusion process. In particular, we present an adaptive method that allows the diffusion model to control the degree of improvement by weather conditions. 

%------------------------------------ Figure 2
%###############################################################################################
\begin{figure}[t]
    \begin{minipage}[b]{1.0\linewidth}
    \centering
    \centerline{\includegraphics[width=12.2cm]{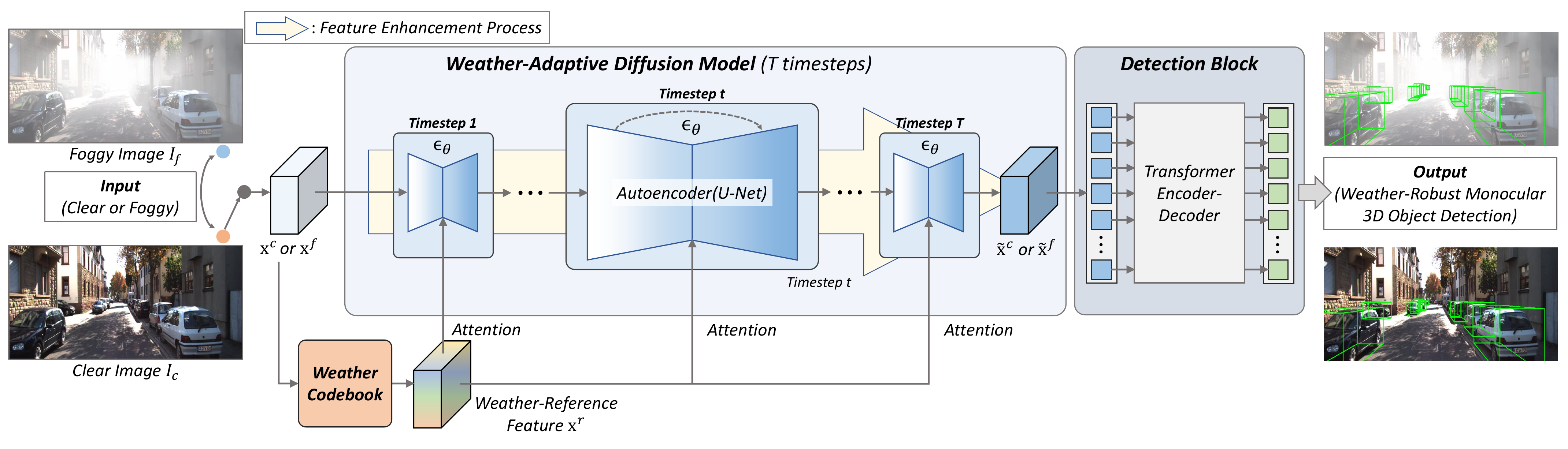}}
    \end{minipage}
    \vspace{-0.6cm}
    \caption{Overview of our MonoWAD in the inference phase. It mainly contains three parts: weather codebook, weather-adaptive diffusion model, and detection block. Through the weather codebook and weather-adaptive diffusion model, our method can maintain robustness against various weather conditions (\textit{i.e.,} clear or foggy).}
    \label{fig:2}
    \vspace{-0.4cm}
\end{figure}
%###############################################################################################

\section{Proposed Method}
Fig. \ref{fig:2} shows the overall framework of the proposed MonoWAD in the inference phase. A backbone network receives an input image (clear image $I_c$ or foggy image $I_f$) to encode the corresponding input feature (input clear feature $x^c$ or input foggy feature $x^f$). It interacts with the weather codebook $\mathcal{Z}$ to generate weather-reference feature $x^r$, indicating the amount of enhancement for the given input feature. Subsequently, the weather-adaptive diffusion model attempts to enhance the input feature over $T$ timesteps to obtain an enhanced feature $\tilde{x}^c$ or $\tilde{x}^f$. Finally, monocular 3D object detection is performed through the detection block. Note that, in the training phase, our MonoWAD use the clear feature $x^c$ to train our diffusion as well as the detection block.

We address two key issues: (1) how to guide weather-reference feature $x^r$ to serve as a reference feature, and (2) how to guide the weather-adaptive diffusion model to effectively enhance feature representations based on weather conditions. Details are in the following subsections.

\vspace{-0.25cm}
\subsection{Weather Codebook}
\label{sec:codebook}

In foggy weather conditions, the overall visual quality of the scene is generally poor, requiring a significant enhancement. Conversely, in clear weather, the amount of improvement is expected to be relatively minimal compared to foggy conditions. Therefore, inspired by \cite{VQ-GAN, VQ-VAE}, we devise a weather codebook $\mathcal{Z}$ to provide the reference knowledge about the weather for appropriate enhancement based on the weather conditions. At this time, as the clear weather contains abundant visual representations, we use it as a reference weather knowledge.

\begin{figure}[t]
	\begin{minipage}[b]{1.0\linewidth}
		\centering
		\centerline{\includegraphics[width=12.5cm]{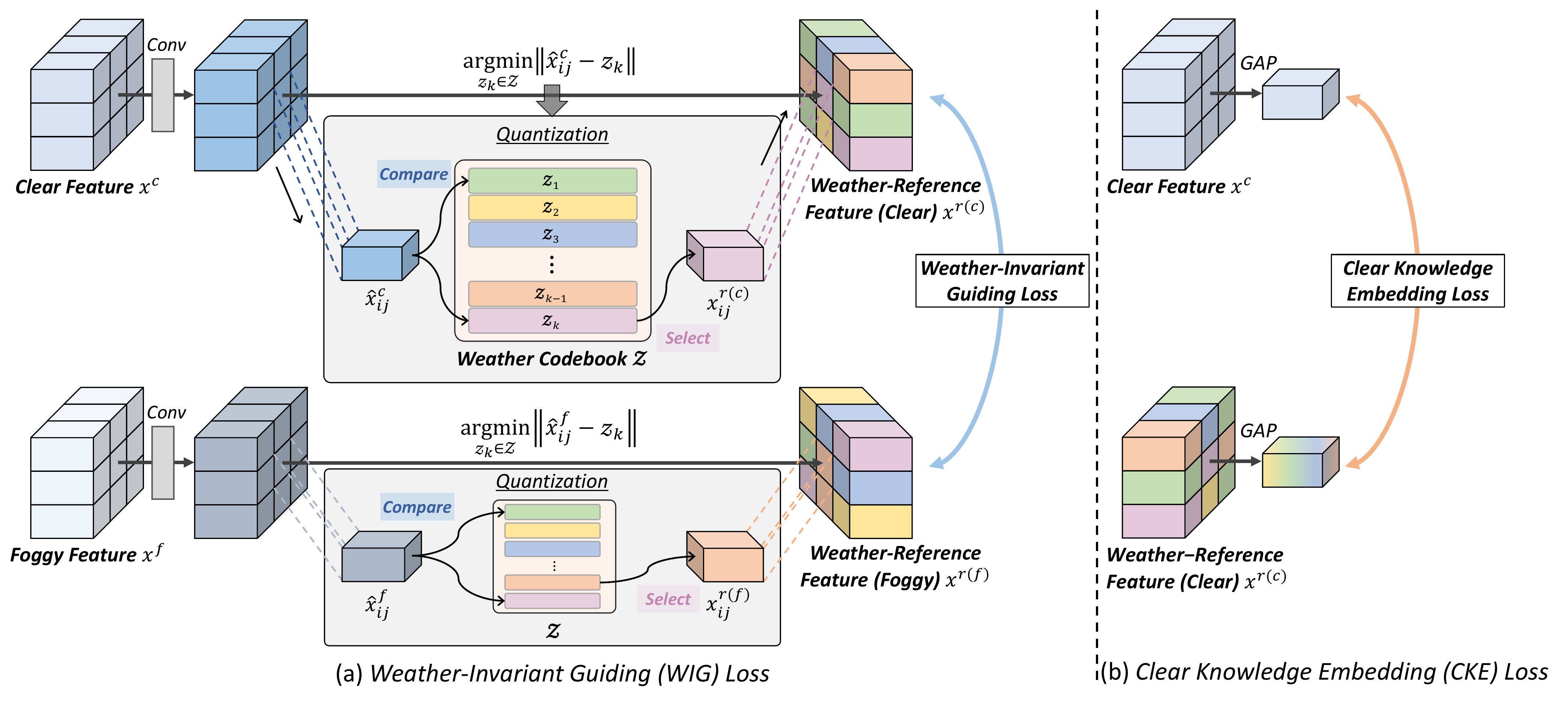}}
	\end{minipage}
    \vspace{-0.75cm}
	\caption{Illustration of the proposed (a) weather-invariant guiding (WIG) loss and (b) clear knowledge embedding (CKE) loss. The clear knowledge recalling (CKR) loss, obtained from combining WIG and CKE, aims to memorize the knowledge of the clear weather and recall the same clear knowledge from the foggy weather.}
    \vspace{-0.5cm}
    \label{fig:codebook}
\end{figure}

As shown in Fig. \ref{fig:codebook}, the reference knowledge embedding procedure involves receiving paired clear-foggy features during the training phase. The weather codebook $\mathcal{Z}$ consists of $K$ learnable slots, denoted as $\mathcal{Z} = \{z_k\}_{k=1}^K(z_k \in \mathbb{R}^{1 \times c})$, where $c$ represents the dimensionality of each slot. The paired clear feature $x^c$ and foggy feature $x^f$ pass through a convolution layer to generate $\hat{x}^c\in \mathbb{R}^{h\times w\times c}$ and $\hat{x}^f\in \mathbb{R}^{h\times w\times c}$ ($w$ denotes width and $h$ indicates height). Each element of the feature denoted as $\hat{x}^{c}_{ij}\in \mathbb{R}^{1 \times c}$ and $\hat{x}^{f}_{ij}\in \mathbb{R}^{1 \times c}$. Subsequently, we obtain weather-reference feature for clear weather, $x^{r(c)}\in \mathbb{R}^{h\times w\times c}$, by conducting element-wise quantization process $\mathbf{q}(\cdot)$, calculated as:
\begin{equation}
  x^{r(c)} = \mathbf{q}(\hat{x}^c) \coloneqq
  \left(\argmin_{z_k \in \mathcal{Z}} \Vert \hat{x}^{c}_{ij} - z_k \Vert\right).
\end{equation}
Utilizing $x^c$ and $x^{r(c)}$, we introduce a clear knowledge embedding (CKE) loss $\mathcal{L}_{cke}$ to guide $x^{r(c)}$ to follow the representation of $x^c$. To this end, we perform global average pooling (GAP) for $x^c$ and $x^{r(c)}$ with softmax, generating $s^c$ and $s^{r(c)}$, respectively. Each element in the vector indicates the probability of the significance of each channel. With $s^c$ and $s^{r(c)}$, we employ KL divergence $D_{KL}(\cdot)$ for $\mathcal{L}_{cke}$ to compare the probability distributions, formulated as:
\begin{equation}
    \mathcal{L}_{cke} = D_{KL}(s^c||s^{r(c)}).
    \label{eq:2}
\end{equation}
Through $\mathcal{L}_{cke}$, the weather codebook $\mathcal{Z}$ can memorize the knowledge of the clear weather, allowing it to effectively reconstruct the clear weather knowledge.

%------------------------------------ Figure 3
%##################################################################################################
%##################################################################################################

Additionally, as the paired clear-foggy images are identical except for weather conditions, the quantization process of the foggy feature $x^f$ with the weather codebook should generate an equivalent weather-reference feature. For obtaining weather-reference feature for foggy $x^{r(f)}$, the element-wise quantization process is also conducted for $\hat{x}^f$ and $\mathcal{Z}$:
\begin{equation}
  x^{r(f)} = \mathbf{q}(\hat{x}^f) \coloneqq
  \left(\argmin_{z_k \in \mathcal{Z}} \Vert \hat{x}^f_{ij} - z_k \Vert\right).
  \label{eq:3}
\end{equation}
Next, we introduce the weather-invariant guiding (WIG) loss $\mathcal{L}_{wig}$ to guide $\mathcal{Z}$ that the weather codebook recalls the same clear knowledge of clear feature for the foggy feature, which can be represented as follows:
\begin{equation}
    \mathcal{L}_{wig}=\left \| x^{r(c)} - x^{r(f)} \right \|_2^2
    \label{eq:4}
.\end{equation}
Finally, the clear knowledge recalling (CKR) loss $\mathcal{L}_{ckr}$ is obtained by adding $\mathcal{L}_{cke}$ and $\mathcal{L}_{wig}$, which is defined as:
\begin{equation}
    \mathcal{L}_{ckr} = \mathcal{L}_{cke} + \mathcal{L}_{wig}.
    \label{eq:5}
\end{equation}

In the training phase, the weight parameters of embedding $K$ slots of weather codebook $\mathcal{Z}$ are initialized randomly and they are updated through Eq. (\ref{eq:5}). In the inference phase, all parameters are fixed to recall clear weather, generating weather-reference features for any weather conditions.

%------------------------------------ Figure 4
%###############################################################################################
\begin{figure*}[t]
    \begin{minipage}[b]{1.0\linewidth}
    \centering
    \centerline{\includegraphics[width=12.0cm]{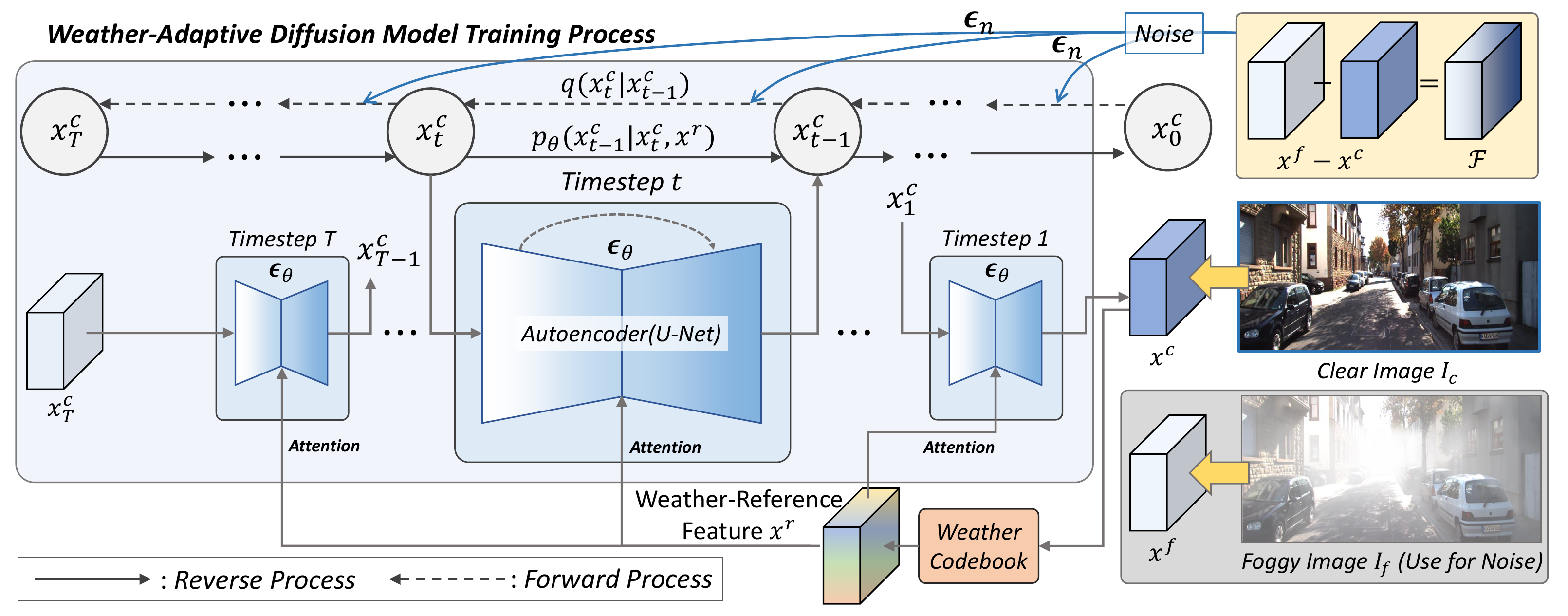}}
    \end{minipage}
    \vspace{-0.75cm}
    \caption{Training process of the weather-adaptive diffusion model, which consists of two processes: (1) Adding fog variant $\boldsymbol{\epsilon}_n$ from input clear feature $x^c$ (forward process) and (2) enhancing representation with weather-reference feature $x^r$ (reverse process).}
    \label{fig:diffusion}
    \vspace{-0.4cm}
\end{figure*}
%###############################################################################################

\subsection{Weather-Adaptive Diffusion Model}
Through Section~\ref{sec:codebook}, we now know that the weather codebook outputs the weather-reference feature $x^{r(c)}$ for the clear weather and $x^{r(f)}$ for the foggy weather through Eq. (\ref{eq:5}). From now on, since our method can receive any input images (clear or foggy), we denote the weather-reference feature as $x^r$.

Fig. \ref{fig:diffusion} shows the training process of the proposed weather-adaptive diffusion model. The key idea of the diffusion model \cite{DDPM, latent-diffusion} is to gradually enhance $x_T$ to $x_0$ with fixed Markov Chain of $T$ timesteps. To this end, the forward and reverse processes are conducted in the training phase, and only the reverse process is used for the inference phase. Motivated by \cite{DDPM, latent-diffusion}, we construct the weather-adaptive diffusion model to enhance representation related to the weather conditions. To this end, unlike traditional diffusion methods \cite{DDPM, latent-diffusion} that adopt the Gaussian noise to the image or latent space, we adopt $\mathcal{F}=x^{f}-x^{c}$, called fog distribution, to guide our diffusion model to be aware of the foggy weather. Ideally, the information contained within $\mathcal{F}$ should include the information of fog, as it represents the difference between the foggy scene and the same scene with clear weather. By doing so, our diffusion model learns the variation in weather by repeatedly adding and removing the fog. Note that, as we take clear feature $x^c$ for $x_0$ to make a reference input for our diffusion model, we newly denote $x_0$ as $x^c_0$.

For the forward process at the $t$-th timestep, $q(x_t^c|x_{t-1}^c)$ takes the previous feature $x_{t-1}^c$ and the noise related to the fog (\textit{i.e.,} $\mathcal{F}$) as inputs to generate $x_{t}^c$. This procedure is repeated over $T$ timesteps, which can be represented as:
\begin{equation}
    q(x^c_t|x^c_{t-1}) = \mathcal{F}(x^c_t;\sqrt{1-\beta_t}x^c_{t-1},\beta_t \bf I),
\end{equation}
\begin{equation}
    q(x^c_1, ..., x^c_T | x^c) = \prod_{t=1}^T q(x^c_t | x^c_{t-1}),
\end{equation}
where $\beta_t$ denotes the variance schedule.

Next, the reverse process at the $t$-th timestep aims to estimate fog variant $\boldsymbol{\epsilon}_n$ using $x_t^c$ to enhance the feature representation of the foggy. To this end, we adopt conditional autoencoder $\boldsymbol{\epsilon}_\theta(x^c_t, t, x^r)$ that receives $x^c_t$ and reference feature $x^r$ from the weather codebook $\mathcal{Z}$. Specifically, $\boldsymbol{\epsilon}_\theta$ estimates mean $\boldsymbol{\mu}_\theta$ and variance $\boldsymbol{\Sigma}_\theta$ of the fog distribution at the $t$-th timestep, denoted as $\tilde{\mathcal{F}}(\cdot)$. The reverse process is also repeated over $T$ timesteps, which can be represented as:
\begin{equation}
    p_\theta(x^c_{t-1}|x^c_t, x^r) = \tilde{\mathcal{F}}(x^c_{t-1};\boldsymbol{\mu}_\theta(x^c_t, t, x^r), \boldsymbol{\Sigma}_\theta(x^c_t, t)),
\end{equation}
\begin{equation}
    p_\theta(x^c, ..., x^c_T) = p(x^c_T)\prod_{t=1}^T p_\theta(x^c_{t-1} | x^c_t, x^r),
\end{equation}
where $p_\theta(x^c_{t-1}|x^c_t, x^r)$ includes cross-attention layer in $\boldsymbol{\epsilon}_\theta$.

The cross-attention layer receives the flattened clear feature $x^c_{t}$ and weather-reference feature $x^{r}$, \textit{i.e.}, $\bar{x}^c_{t}$ and  $\bar{x}^r$, respectively. The similarity between $\bar{x}^{r}$ and $\bar{x}^{c}_t$ is calculated in the cross-attention layer to transfer the enhancement to $\bar{x}^{c}_t$, which can be formulated as:
\begin{equation}
    \text{Attention}(Q, K, V) = \text{softmax}\left(\frac{QK^T}{\sqrt{d}}\right)V,
\end{equation}
where $Q=W_i^q\cdot\bar{x}^c_{t}, K=W_i^k\cdot\bar{x}^{r}, V=W_i^v\cdot\bar{x}^{r}$ with learnable parameters $W_i^q, W_i^k, W_i^v$.

To ensure that the estimated fog variant $\boldsymbol{\epsilon}_\theta$ is similar to the fog variant $\boldsymbol{\epsilon}_n$ applied to make $x^c_t$ from the input clear feature $x^c$, we propose a weather-adaptive enhancement loss $\mathcal{L}_{wae}$, which is formulated as:
\begin{equation}
    \mathcal{L}_{wae} = \mathbb{E}_{x^c, \boldsymbol{\epsilon}_n \sim \mathcal{F}, t }\Big[ \Vert \boldsymbol{\epsilon}_n - \boldsymbol{\epsilon}_\theta(x^c_t, t, x^r) \Vert_{2}^{2}\Big].
\end{equation}

\noindent With $\mathcal{L}_{wae}$, $\boldsymbol{\epsilon}_\theta$ can estimate the fog variant by leveraging the fog distribution as noise for our diffusion model through multiple forward/reverse processes. Additionally, the cross-attention layer within $\boldsymbol{\epsilon}_\theta$ dynamically enhances the feature representation based on combining knowledge of the input feature (foggy or clear) and the weather-reference feature. Since our diffusion model learns the degree of improvement needed corresponding to weather conditions, it can improve the representation of any input, whether clear or foggy in inference phase. This leads our monocular 3D object detector can handle both clear and foggy input images, resulting in weather-robust detection.

\vspace{-0.3cm}
\subsection{Total Loss Function}
The total loss function of our MonoWAD is represented as follows:
\begin{equation}
    \mathcal{L}_{Total} = \mathcal{L}_{OD} + \lambda_1\mathcal{L}_{ckr} + \lambda_2\mathcal{L}_{wae},
\end{equation}
where $\lambda_1$ and $\lambda_2$ denote balancing hyper-parameters. In our experiments, we set $\lambda_1=\lambda_2 = 1$. $\mathcal{L}_{OD}$ is the detection loss for 3D object detection \cite{MonoDTR,MonoDETR}. It includes loss functions of classification, regression, and depth loss, that are similar to prior works \cite{D4LCN,M3DSSD,MonoDTR}. The overall weight parameters are updated through $\mathcal{L}_{Total}$.

\section{Experiments}

\label{sec:exper}
\subsection{Dataset and Evaluation Metrics}
\subsubsection{Datasets.}
We utilize the KITTI 3D object detection dataset \cite{KITTI}, the most widely adopted for 3D object detection. It contains 7,481 training images and 7,518 test images under clear weather conditions. Due to unavailable ground-truth annotations for test images and limited evaluation on the test server, we follow \cite{Monocular_3d} by splitting 3,712 training images and 3,769 validation images. In addition, as our work requires paired images to learn about weather changes, we generate foggy images from all images in the KITTI dataset, called foggy KITTI, that emulate the foggy scene. Following the protocols of \cite{Foggy_Cityscapes, SeeingThroughFog}, we generate foggy images based on object distances using depth maps estimated by DORN \cite{DORN}. Please refer to the supplementary materials for details.

In addition, as our work is focused on robust monocular 3D object detection in various weather conditions, we further adopt the Virtual KITTI dataset \cite{Virtual_KITTI1, Virtual_KITTI2}, which contains photo-realistic synthetic images under various weather conditions (\textit{e.g.,} foggy, rainy, sunset). It is associated with the original real-world KITTI dataset, which has 3D annotations for each weather.

\vspace{-0.4cm}
\subsubsection{Evaluation Metrics.}

We adopt average precision in both 3D detection (AP$_{3D}$) and bird-eye view detection (AP$_{BEV}$) under three difficulty levels (`Easy', `Moderate', `Hard') according to size, occlusion, and truncation. Following \cite{Disentangling_monocular}, we use 40 recall position metric AP$_{40}$ and report scores for the car category under the IoU threshold 0.7 for the KITTI dataset and 0.5 for the Virtual KITTI dataset.
\vspace{-0.5cm}
\subsection{Implementation Details}
We use DLA-102 \cite{DLA} as our backbone and adopt the transformer architecture of \cite{MonoDTR} for the detection block. We train MonoWAD on a single RTX 4090 GPU with a batch size of 4 over 120 epochs using Adam optimizer \cite{ADAM} (initial learning rate of $10^{-4}$). For the weather codebook, we utilize embedding slot $K=4096$ and set the dimension of each slot $D=256$. We set 4 heads for cross-attention of the weather-adaptive diffusion model, and the timesteps for the forward and reverse process of diffusion default to 15. The number of channels for the diffusion output features is $C=256$. As recent monocular 3D object detectors \cite{GUPNet,DID-M3D,MonoGround,MonoDTR,MonoDETR} have not been explored under adverse weather conditions, we implemented them with available official source code to faithfully reproduce them. 

%------------------------------------ Table 1
%##################################################################################################
\begin{table}[tb]
    \caption{Detection results of car category on KITTI validation set under foggy weather and clear weather conditions. The results of the state-of-the-art methods under foggy weather are obtained through our reproduction with the official source code. \textbf{Bold}/\underline{underlined} fonts indicate the best/second-best results.}
    \vspace{-0.63cm}
    \renewcommand{\tabcolsep}{1.8mm}
	\centering
	\begin{center} 
		\resizebox{0.999\linewidth}{!}
		{
			\begin{tabular}{c ccc ccc ccc}
				\Xhline{3\arrayrulewidth}
                \rule{0pt}{10.5pt} \multirow{2}{*}[-0.4em]{\bf Method} & \multicolumn{3}{c}{\bf Foggy (\bf AP$_{3D}$)} & \multicolumn{3}{c}{\bf Clear (\bf AP$_{3D}$)} & \multicolumn{3}{c}{\multirow{1}{*}{\bf Average}}
                \\\cmidrule(lr){2-4}\cmidrule(lr){5-7}\cmidrule(lr){8-10}
				
				& \bf Easy & \bf Mod. & \bf Hard & \bf Easy & \bf Mod. & \bf Hard & \bf Easy & \bf Mod. & \bf Hard\\\hline
				\rule{0pt}{10.5pt}
				GUPNet \cite{GUPNet} (ICCV'21) 
				& 2.74 & 2.19 & 2.16 & 22.76 & 16.46 & 13.72 & 12.75 & 9.33  & 7.94 \\
				DID-M3D \cite{DID-M3D} (ECCV'22) 
				& 1.15 & 0.61 & 0.64 & 22.98 & 16.12 & 14.03 & 12.07 & 8.37 & 7.34\\
				MonoGround \cite{MonoGround} (CVPR'22) 
				& 0.00 & 0.00 & 0.06 & 25.24 & 18.69 & 15.58 & 12.62 & 9.35 & 7.82\\
				MonoDTR\cite{MonoDTR} (CVPR'22) 
				& \underline{16.89} & \underline{11.86} & \underline{9.87} & 24.52 & 18.57 & 15.51 & \underline{20.71} & \underline{15.22} & \underline{12.69} \\
				MonoDETR \cite{MonoDETR} (ICCV'23) 
				& 7.40 & 5.74 & 4.53 & \underline{28.84} & \underline{20.61} & \underline{16.38} & 18.12 & 13.18 & 10.46 \\
				\cdashline{1-10}
				\rule{0pt}{10.8pt} \bf MonoWAD (Ours)
				& \bf 27.17 & \bf 19.57 & \bf 16.21 & \bf 29.10 & \bf 21.08 & \bf 17.73 & \bf 28.14 & \bf 20.33 & \bf 16.97\\
				\Xhline{3\arrayrulewidth}
			\end{tabular}
		}
        \vspace{-0.4cm}
	\label{table:valid}
	\end{center}
\end{table}
%##################################################################################################

%------------------------------------ Table 2
%##################################################################################################
\begin{table}[t]
    \caption{Detection results of car category on KITTI test set under foggy weather. \textbf{Bold}/\underline{underlined} fonts indicate the best/second-best results.}
    \vspace{-0.6cm}
    \renewcommand{\tabcolsep}{2.5mm}
	\centering
	\begin{center}
		\resizebox{0.85\linewidth}{!}
		{
			\begin{tabular}{c ccc ccc}
				\Xhline{3\arrayrulewidth}
				\rule{0pt}{10pt} \multirow{2}{*}[-0.1em]{\bf Method}  & \multicolumn{3}{c}{\bf AP$_{3D}$} & \multicolumn{3}{c}{\bf AP$_{BEV}$}
				\\\cmidrule(lr){2-4}\cmidrule(lr){5-7}
				& \bf Easy & \bf Mod. & \bf Hard & \bf Easy & \bf Mod. & \bf Hard \\\hline
				\rule{0pt}{9.5pt}
				GUPNet \cite{GUPNet} (ICCV'21)
				& 3.01 & 2.42 & 1.13 & 4.90 & 3.02 & 2.91\\
				DID-M3D \cite{DID-M3D} (ECCV'22)
				& 3.10 & 2.39 & 2.19 & 5.34 & 3.01 & 2.91\\
				MonoGround \cite{MonoGround} (CVPR'22)
				& 0.14 & 0.20 & 0.22 & 0.23 & 0.38 & 0.39\\
				MonoDTR \cite{MonoDTR} (CVPR'22)
				& \underline{11.07} & \underline{7.41} & \underline{5.26} & \underline{15.76} & \underline{10.15} & \underline{7.53}\\
				MonoDETR \cite{MonoDETR} (ICCV'23)
				& 9.33 & 5.54 & 4.06 & 13.15 & 8.06 & 6.30\\
				\cdashline{1-7}
				\rule{0pt}{10.8pt} \bf MonoWAD (Ours) 
				& \bf 19.75 & \bf 13.32 & \bf 11.04 & \bf 27.95 & \bf 19.06 & \bf 15.61\\
				\Xhline{3\arrayrulewidth}
			\end{tabular}
		}
        \vspace{-0.7cm}
		\label{table:test}
	\end{center}
\end{table}
%##################################################################################################

\vspace{-0.2cm}
\subsection{Comparison}

\subsubsection{Results on KITTI 3D Dataset.}
We compared MonoWAD with state-of-the-art monocular 3D object detectors \cite{GUPNet, DID-M3D, MonoGround, MonoDTR, MonoDETR} that do not use additional data (\textit{e.g.,} depth maps or LiDAR) during inference on the KITTI and foggy KITTI validation sets. Table \ref{table:valid} shows the $AP_{3D}$ results, and $AP_{BEV}$ results are in the supplementary materials. While recent methods have shown improved performance in clear weather, their performance drops in foggy weather, limiting their applicability in real-world applications (\textit{e.g.,} autonomous driving and robotics). In contrast, MonoWAD showed stable 3D detection performance under both foggy and clear weather. Since our weather codebook learns knowledge about clear weather and the weather-adaptive diffusion model can enhance the feature representation of the input images under both clear and foggy weather, MonoWAD shows a more weather-robust detection performance than that of the existing methods. Also, to explore the weather robustness of MonoWAD, we conducted experiments with mixed foggy and clear weather conditions at various ratios (please see the supplementary materials).

We further compared 3D detection performances on the foggy KITTI test set (Table \ref{table:test}). Similar to Table \ref{table:valid}, existing monocular 3D object detectors show lower performance under foggy weather. Through the results, our method shows the weather-robust monocular 3D performance under foggy and clear weather.

%------------------------------------ Table 3
%##################################################################################################
\begin{table}[tb]
    \caption{Detection results of car category on Virtual KITTI under foggy, rainy, sunset conditions. \textbf{Bold}/\underline{underlined} fonts indicate the best/second-best results.}
    \vspace{-0.6cm}
    \renewcommand{\tabcolsep}{1.8mm}
	\centering
	\begin{center} 
		\resizebox{0.99\linewidth}{!}
		{
			\begin{tabular}{c ccc ccc ccc}
				\Xhline{3\arrayrulewidth}
                \rule{0pt}{10.5pt} \multirow{2}{*}[-0.4em]{\bf Method} & \multicolumn{3}{c}{\bf Foggy (\bf AP$_{3D}$)} & \multicolumn{3}{c}{\bf Rainy (\bf AP$_{3D}$)} & \multicolumn{3}{c}{\bf Sunset (\bf AP$_{3D}$)}
                \\\cmidrule(lr){2-4}\cmidrule(lr){5-7}\cmidrule(lr){8-10}
				
				& \bf Easy & \bf Mod. & \bf Hard & \bf Easy & \bf Mod. & \bf Hard & \bf Easy & \bf Mod. & \bf Hard\\\hline
				\rule{0pt}{10.5pt}
                GUPNet \cite{GUPNet} (ICCV'21)
				& 1.76 & 1.57 & 1.57 & 2.34 & 1.24 & 1.21 & 2.77 & 1.64 & 1.65 \\
                DID-M3D \cite{DID-M3D} (ECCV'22)
				& 0.91 & 0.39 & 0.39 & 0.40 & 0.13 & 0.13 & 0.34 & 0.10 & 0.10 \\
                MonoGround \cite{MonoGround} (CVPR'22)
				& 0.29 & 0.30 & 0.25 & 5.49 & 2.82 & 2.77 & 7.68 & 4.24 & 4.20 \\
				MonoDTR \cite{MonoDTR} (CVPR'22)
				& \underline{8.79} & \underline{5.75} & \underline{5.72} & \underline{11.73} & \underline{6.25} & \underline{6.74} & \underline{9.86} & \underline{5.42} & \underline{5.42} \\
                MonoDETR \cite{MonoDETR} (ICCV'23) 
			  & 4.50 & 2.99 & 2.96 & 6.61 & 3.46 & 3.42 & 7.08 & 4.17 & 4.16 \\
				\cdashline{1-10}
				\rule{0pt}{10.8pt} \bf MonoWAD (Ours)
				& \bf 13.33 & \bf 8.56 & \bf 8.50 & \bf 14.12 & \bf 8.33 & \bf 8.24 & \bf 13.38 & \bf 7.89 & \bf 7.80\\
				\Xhline{3\arrayrulewidth}
			\end{tabular}
		}
        \vspace{-0.5cm}
	\label{table:virtual}
	\end{center}
\end{table}
%##################################################################################################

\vspace{-0.3cm}
\subsubsection{Results on Virtual KITTI Dataset.}
We also conducted experiments on the Virtual KITTI dataset to validate the generalization ability of our method across various weather conditions (\textit{i.e.,} foggy, rainy, sunset). The results are shown in Table \ref{table:virtual}. MonoWAD also shows the highest performance in foggy weather. Moreover, it outperforms the existing method in the rainy and sunset conditions. The results demonstrate that MonoWAD is robust and insensitive to various weather conditions that can be faced in real-world scenarios, not just clear weather.

%------------------------------------ Table 4
%##################################################################################################
\begin{table}[t!]
    \caption{Effect of the proposed method on KITTI validation set for car category. \textbf{WC} denotes our weather codebook, \textbf{WAD} indicates our weather-adaptive diffusion model.}
    \vspace{-0.7cm}
     \renewcommand{\tabcolsep}{1.0mm}
	\centering
	\begin{center}
		\resizebox{0.78\linewidth}{!}
		{
			\begin{tabular}{c c c ccc ccc}
				\Xhline{3\arrayrulewidth}
				\multirow{2}{*}[-0.4em]{\bf Method} & \multirow{2}{*}[-0.4em]{\bf WC} & \multirow{2}{*}[-0.4em]{\bf WAD} & 
                \multicolumn{3}{c}{\bf Foggy (AP$_{3D}$)} & \multicolumn{3}{c}{\bf Clear (AP$_{3D}$)}
                \\\cmidrule(lr){4-7}\cmidrule(lr){7-9}
                
				& & & \bf Easy & \bf Mod. & \bf Hard & \bf Easy & \bf Mod. & \bf Hard \\\hline
				\rule{0pt}{10.5pt}
				Baseline & - & -
				& 13.75 & 9.61 & 8.10 & 22.63 & 17.16 & 14.28 \\\cdashline{1-9}
				\rule{0pt}{10.5pt} \multirow{2}{*}{\bf MonoWAD (Ours)} & - & \cmark 
				& 25.62 & 18.66 & 15.56 & 26.34 & 19.17 & 16.15 \\
				
				& \cmark & \cmark
				& \bf 27.17 & \bf 19.57 & \bf 16.21 & \bf 29.10 & \bf 21.08 & \bf 17.73 \\
				\Xhline{3\arrayrulewidth}
			\end{tabular}
		}
        \vspace{-0.45cm}
		\label{table:Effect_method}
	\end{center}
\end{table}
%##################################################################################################

%------------------------------------ Table 5
%##################################################################################################
\begin{table}[t!]
    \caption{Detection results on KITTI validation set by changing diffusion timestep $\textit{T}$.}
    \vspace{-0.7cm}
    \renewcommand{\tabcolsep}{3.5mm}
	\centering
	\begin{center}
		\resizebox{0.8\linewidth}{!}
		{
			\begin{tabular}{c ccc ccc}
				\Xhline{3\arrayrulewidth}
				\rule{0pt}{10pt} \multirow{2}{*}[-0.4em]{\bf Timestep $\textit{T}$} & \multicolumn{3}{c}{\bf Foggy (AP$_{3D}$)} & \multicolumn{3}{c}{\bf Clear (AP$_{3D}$)}
                \\\cmidrule(lr){2-4}\cmidrule(lr){5-7}
                
				& \bf Easy & \bf Mod. & \bf Hard & \bf Easy & \bf Mod. & \bf Hard \\\hline
				\rule{0pt}{9.5pt} -
				& 13.75 & 9.61 & 8.10 & 22.63 & 17.16 & 14.28 \\\cdashline{1-7}
                \rule{0pt}{11.5pt} 5
				& 23.57 & 17.91 & 14.96 & 26.03 & 19.21 & 16.01 \\
				\rule{0pt}{5pt} 10
				& 25.28 & 18.49 & 15.42 & 26.79 & 19.90 & 16.78\\
				\rule{0pt}{5pt} \textbf{15}
				& \bf 27.17 & \bf 19.57 & \bf 16.21 & \bf 29.10 & \bf 21.08 & \bf 17.73 \\
				\rule{0pt}{5pt} 20
				& 24.54 & 18.29 & 15.10 & 24.85 & 18.54 & 15.34 \\
                
				\Xhline{3\arrayrulewidth}
			\end{tabular}
		}
            \vspace{-0.75cm}
		\label{table:timestep}
	\end{center}
\end{table}
%###############################################################################################

\subsection{Ablation Studies}
We conducted ablation studies to examine (1) the effect of each proposed component (\textit{i.e.,} weather codebook and weather-adaptive diffusion model) and (2) the effect of the weather-adaptive diffusion model by varying the timestep \textit{T}. These experiments were performed on the KITTI and foggy KITTI 3D validation sets.

%------------------------------------ Table 6
%##################################################################################################
\begin{table}[tb]
    \caption{Performance comparison on KITTI validation set under foggy and clear weather conditions. We compare our MonoWAD with MonoDTR \cite{MonoDTR}, which demonstrates superior performance in foggy weather using state-of-the-art dehazing methods.}
    \vspace{-0.65cm}
    \renewcommand{\tabcolsep}{0.6mm}
	\centering
	\begin{center} 
		\resizebox{0.99\linewidth}{!}
		{
			\begin{tabular}{c c ccc ccc ccc}
				\Xhline{3\arrayrulewidth}
                \rule{0pt}{10.5pt} \multirow{2}{*}[-0.4em]{\bf Method} & \multirow{2}{*}[-0.4em]{\bf \makecell{Dehazing\\Method}} & \multicolumn{3}{c}{\bf Foggy (AP$_{3D}$)} & \multicolumn{3}{c}{\bf Clear (AP$_{3D}$)} & \multicolumn{3}{c}{\multirow{1}{*}{\bf Average}}
                \\\cmidrule(lr){3-5}\cmidrule(lr){6-8}\cmidrule(lr){9-11}
				
				& & \bf Easy & \bf Mod. & \bf Hard & \bf Easy & \bf Mod. & \bf Hard & \bf Easy & \bf Mod. & \bf Hard\\\hline
				\rule{0pt}{10.5pt}
				MonoDTR($\mathcal{B}$)\cite{MonoDTR} (CVPR'22) &
				- & 16.89 & 11.86 & 9.87 & 24.52 & 18.57 & 15.51 & 20.71 & 15.22 & 12.69 \\\cdashline{1-11}

                \rule{0pt}{10.5pt}
                $\mathcal{B}$ + RIDCP \cite{RIDCP} (CVPR'23) &
				Image-level & 17.23 & 12.41 & 10.44 & 24.02 & 17.89 & 14.78 & 20.63 & 15.15 & 12.61 \\
    
				$\mathcal{B}$ + DENet \cite{DENet} (ACCV'22) &
				Feature-level & 22.35 & 17.44 & 14.47 & 7.10 & 5.70 & 4.53 & 14.73 & 11.57 & 9.50 \\
                $\mathcal{B}$ + Yang \textit{et al.} \cite{DA_foggy} (ACCV'22) &
			    Feature-level & 22.87 & 15.21 & 12.17 & 17.96 & 13.10 & 10.64 & 20.42 & 14.16 & 11.41 \\
				\cdashline{1-11}
				\rule{0pt}{10.8pt} \bf MonoWAD (Ours) &
				- & \bf 27.17 & \bf 19.57 & \bf 16.21 & \bf 29.10 & \bf 21.08 & \bf 17.73 & \bf 28.14 & \bf 20.33 & \bf 16.97\\
				\Xhline{3\arrayrulewidth}
			\end{tabular}
		}
        \vspace{-0.35cm}
	\label{table:dehaze}
	\end{center}
\end{table}
%##################################################################################################

%------------------------------------ Table 8
%##################################################################################################
\begin{table}[t]
    \caption{Comparison of diffusion models on KITTI validation set for car category: $\mathcal{B}$ is baseline detection block (transformer encoder-decoder), \textbf{CDM} (Conditional Diffusion Model \cite{latent-diffusion}), \textbf{WC} (weather codebook), and \textbf{WAD} (weather adaptive diffusion model).}
    \vspace{-0.6cm}
    \renewcommand{\tabcolsep}{2.6mm}
	\centering
	\begin{center}
		\resizebox{0.8\linewidth}{!}
		{
			\begin{tabular}{c ccc ccc}
				\Xhline{3\arrayrulewidth}
				\rule{0pt}{10pt} \multirow{2}{*}[-0.4em]{\bf Method} & \multicolumn{3}{c}{\bf Foggy (AP$_{3D}$)} & \multicolumn{3}{c}{\bf Clear (AP$_{3D}$)}
                \\\cmidrule(lr){2-4}\cmidrule(lr){5-7}
                
				& \bf Easy & \bf Mod. & \bf Hard & \bf Easy & \bf Mod. & \bf Hard \\\hline
				\rule{0pt}{10.5pt}
				$\mathcal{B}$+DDPM \cite{DDPM}
				& 5.32 & 3.84 & 2.77 & 18.31 & 12.71 & 10.20 \\
				$\mathcal{B}$+CDM \cite{latent-diffusion}
				& 2.74 & 2.10 & 2.00 & 20.50 & 14.51 & 11.63 \\
				$\mathcal{B}$+WC+CDM \cite{latent-diffusion}
				& 17.51 & 12.74 & 10.40 & 21.05 & 15.11 & 12.54 \\
                $\mathcal{B}$+WAD
				& 25.62 & 18.66 & 15.56 & 26.34 & 19.17 & 16.15 \\
				\cdashline{1-7}
				\rule{0pt}{11pt} \bf \makecell{MonoWAD \\$(\mathcal{B}$+WC+WAD)}
				& \bf 27.17 & \bf 19.57 & \bf 16.21 & \bf 29.10 & \bf 21.08 & \bf 17.73 \\
				\Xhline{3\arrayrulewidth}
			\end{tabular}
		}
        \vspace{-0.7cm}
		\label{table:diffusion}
	\end{center}
\end{table}
%##################################################################################################

%------------------------------------ Figure 5
%###############################################################################################
\begin{figure*}[t]
    \begin{minipage}[b]{1.0\linewidth}
    \centering
    \centerline{\includegraphics[width=12.0cm]{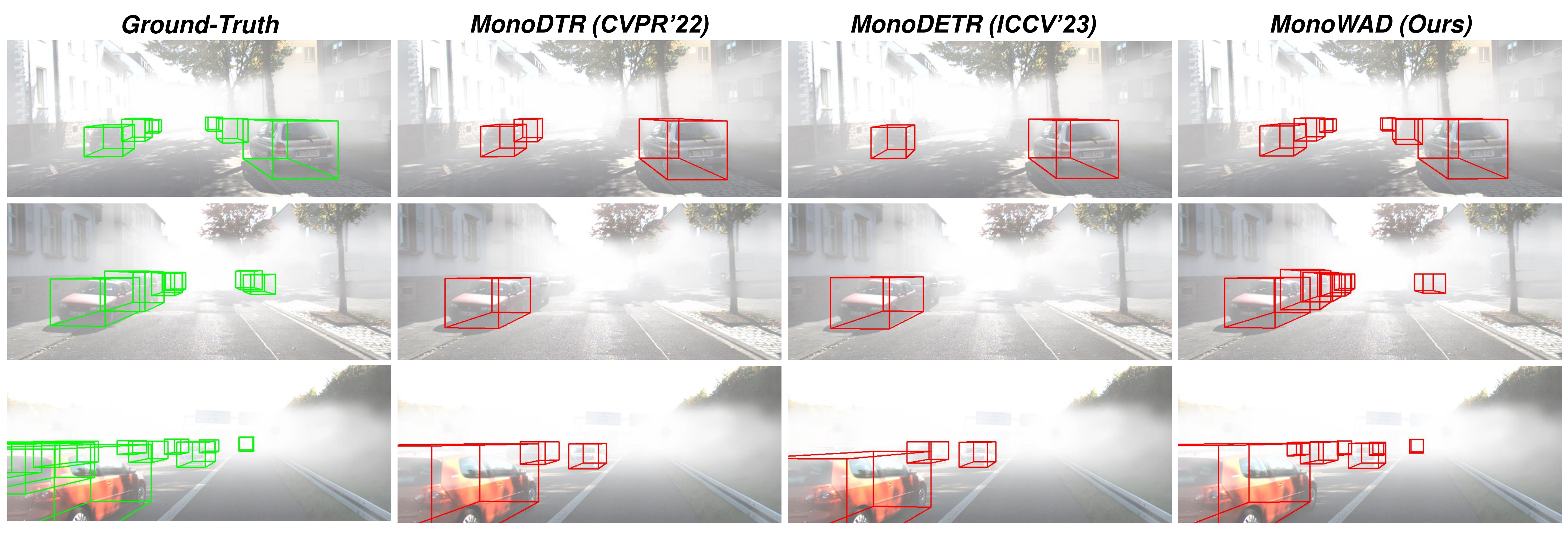}}
    \end{minipage}
    \vspace{-0.8cm}
    \caption{Comparison of 3D detection examples (green: ground-truth, red: predicted 3D bounding-box) between our MonoWAD and two detectors, MonoDTR \cite{MonoDTR} and MonoDETR \cite{MonoDETR}, that show the most improved performances among existing methods.}
    \label{fig:vis}
    \vspace{-0.15cm}
\end{figure*}
%###############################################################################################

%------------------------------------ Figure 6
%###############################################################################################
\begin{figure*}[t]
    \begin{minipage}[b]{1.0\linewidth}
    \centering
    \centerline{\includegraphics[width=12.5cm]{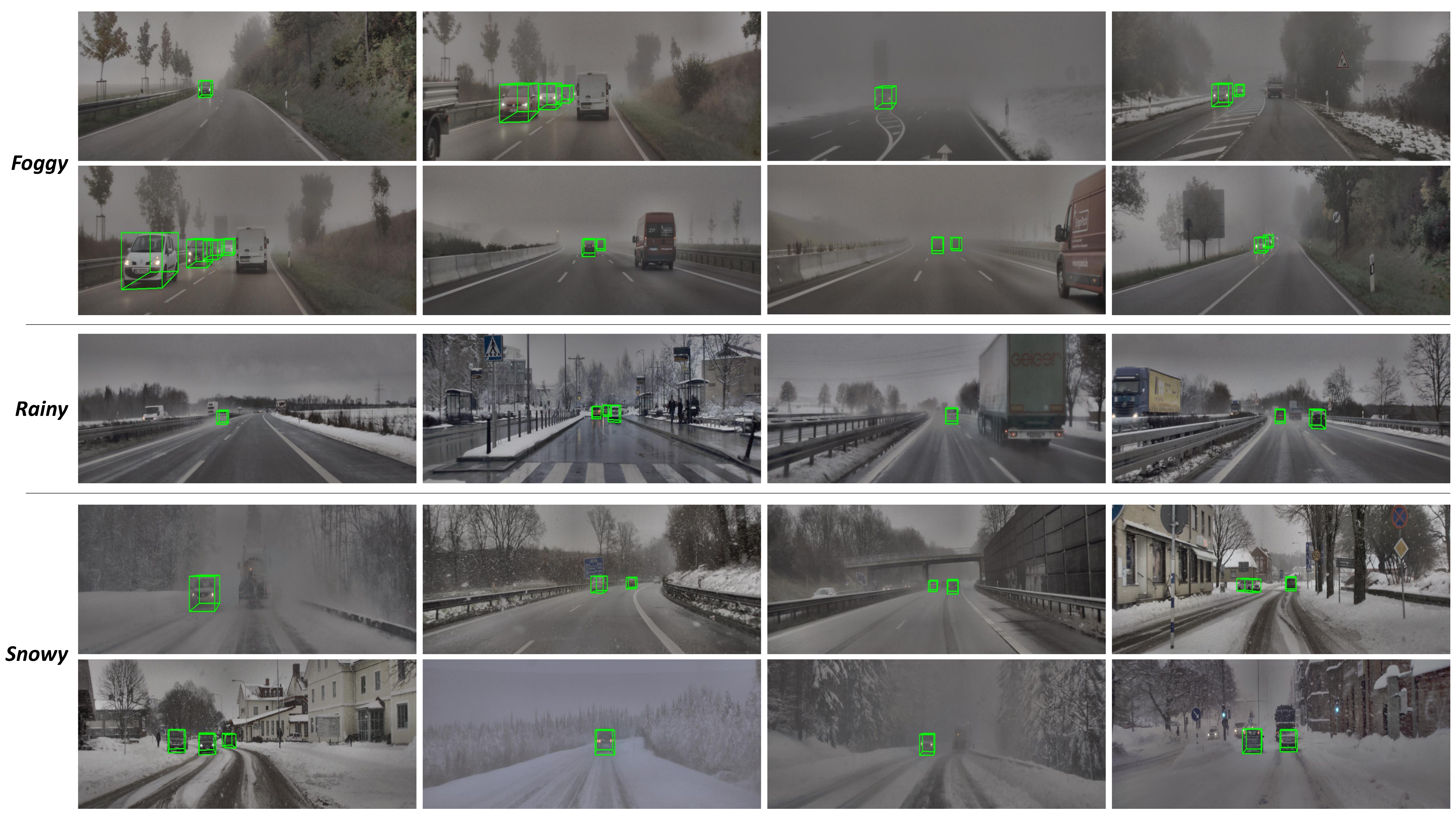}}
    \end{minipage}
    \vspace{-0.8cm}
    \caption{3D detection results on real-world images of various weather conditions.}
    \label{fig:real}
    \vspace{-0.5cm}
\end{figure*}
%###############################################################################################

%------------------------------------ Figure 7
%###############################################################################################
\begin{figure*}[t]
    \begin{minipage}[b]{1.0\linewidth}
    \centering
    \centerline{\includegraphics[width=11.0cm]{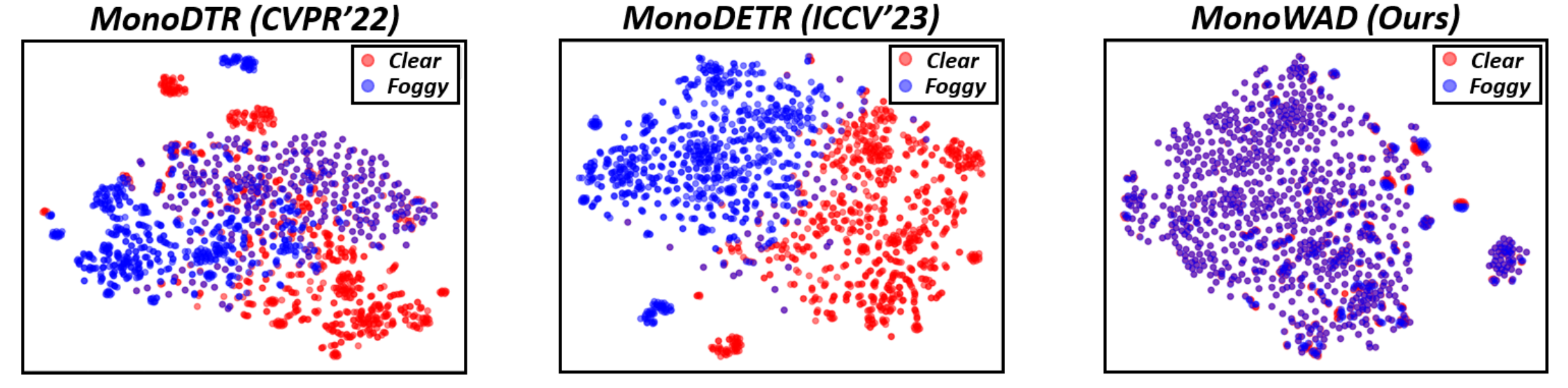}}
    \end{minipage}
    \vspace{-0.8cm}
    \caption{t-SNE visualization results (Red: Clear, Blue: Foggy).}
    \label{fig:tsne}
    \vspace{-0.4cm}
\end{figure*}
%###############################################################################################

\subsubsection{Effect of the Proposed Modules.}

The results regarding the effectiveness of the proposed weather codebook and weather-adaptive diffusion model are presented in Table \ref{table:Effect_method}. Since our weather-adaptive diffusion model is designed to enhance feature representation, it alone has shown significant performance improvement. When the weather codebook is additionally considered, the weather-adaptive diffusion model can leverage the knowledge of the weather-reference feature for clear weather, leading to enhanced performance. It makes MonoWAD can be robust to various weather conditions for monocular 3D object detection.

\vspace{-0.3cm}
\subsubsection{Effect of Timestep $\textit{\textbf{T}}$.}

We also conduct experiments by varying timestep $T$ of the weather-adaptive diffusion model. Timestep is the number of steps for the forward and reverse process of the diffusion model. Table \ref{table:timestep} indicates that the highest monocular 3D detection performance is achieved when $T=15$ under both clear and foggy weather conditions. Our MonoWAD consistently outperforms the baseline and other existing methods across all timesteps.

\vspace{-0.2cm}
\subsection{Discussions}
\subsubsection{Comparison with Dehazing Methods.}

We investigate the weather-robustness of our method for monocular 3D object detection compared to the other monocular detectors using dehazing methods. To this end, we compared MonoWAD with MonoDTR \cite{MonoDTR}, which shows the best performance in foggy weather through the application of state-of-the-art image-level and feature-level dehazing methods \cite{RIDCP, DENet, DA_foggy}. The results are shown in Table \ref{table:dehaze}. Recent dehazing methods are primarily focused on specific weather conditions, such as foggy weather. While they have shown some improvement under foggy conditions, they exhibit reduced performances under clear weather conditions. In contrast, our MonoWAD shows robust performance across both foggy and clear weather conditions.

\vspace{-0.3cm}
\subsubsection{Effect of Weather-Adaptive Diffusion Model.}

Table \ref{table:diffusion} shows the effectiveness of MonoWAD with existing diffusion models \cite{DDPM, latent-diffusion} on the KITTI and foggy KITTI validation sets. Existing methods adopt Gaussian noise for forward and reverse processes, but they can not fully understand about weather. In contrast, our weather-adaptive diffusion understands weather variances, allowing our MonoWAD to surpass existing methods in clear and foggy weather.

\subsection{Visualization Results}
\subsubsection{Results on KITTI Dataset.}

We visualize several 3D detection results on the KITTI 3D dataset, comparing MonoWAD with MonoDTR and MonoDETR, which exhibit the highest performances among existing methods under foggy condition (Fig. \ref{fig:vis}). Existing methods struggle to detect objects obscured by fog, indicating limitations in detecting only fully visible objects. In contrast, even in dense fog, MonoWAD effectively detects both close and fog-obscured objects with the aid of the weather codebook and weather-adaptive diffusion model.
\vspace{-0.3cm}
\subsubsection{Results on Real-World Images.}

We further visualize 3D detection results on real-world images from the Seeing Through Fog dataset\cite{SeeingThroughFog} under various weather conditions (\textit{i.e.,} foggy, rainy, snowy). In Fig. \ref{fig:real}, MonoWAD shows robust detection under diverse weather conditions. This demonstrates that our proposed method maintains weather-robustness even in real-world scenarios by dynamically enhancing the input scenarios.
\vspace{-0.3cm}
\subsubsection{t-SNE Visualization.}
We conducted t-SNE visualization to analyze feature representations of MonoDTR, MonoDETR, and our MonoWAD on the KITTI and foggy KITTI validation set. As depicted in Fig. \ref{fig:tsne}, the existing methods exhibit distinct feature representations for foggy and clear weather conditions. In contrast, MonoWAD, leveraging weather-robust feature learning from the weather codebook and weather-adaptive diffusion model, demonstrates similar feature representations for both clear and foggy weather conditions.
\vspace{-0.12cm}
\subsection{Limitations}
The experimental results show the weather-robustness of our method. However, due to the iterative nature of the diffusion model, our method shows 144ms/image at timestep $T=15$ (110ms/image at $T=5$), slower than the latest work, MonoDETR (38ms/image). Moreover, our method dynamically enhances the representation of the input feature based on the weather-reference feature and weather difference which needs paired images. Thus, exploring a method to achieve faster processing speeds while maintaining weather-robust performance without paired images could be an interesting direction for our future work.
\vspace{-0.2cm}
\section{Conclusion} 
We proposed MonoWAD, a novel weather-robust monocular 3D object detector to handle various weather conditions. Addressing challenges in applying existing monocular 3D object detectors to real-world scenarios with various weather, we design a weather codebook with clear knowledge recalling loss to memorize the knowledge of the clear weather and to generate a weather-reference feature from both clear and foggy features. Also, we design a weather-adaptive diffusion model with weather-adaptive enhancement loss to enhance feature representation according to the weather conditions. As a result, our MonoWAD can detect objects occluded by fog and perform well in clear weather.

\section*{Acknowledgements} This work was supported by the NRF grant funded by the Korea government (MSIT) (No. RS-2023-00252391), and by IITP grant funded by the Korea government (MSIT) (No. RS-2022-00155911: Artificial Intelligence Convergence Innovation Human Resources Development (Kyung Hee University), IITP-2023-RS-2023-00266615: Convergence Security Core Talent Training Business Support Program, No. 2022-0-00124: Development of Artificial Intelligence Technology for Self-Improving Competency-Aware Learning Capabilities).

\bibliographystyle{splncs04}
\bibliography{main}
\clearpage

\renewcommand{\thesection}{\Alph{section}}
\makeatletter
\def\@seccntformat#1{\Alph{section}.\hspace{0.5em}}
\def\section{\@startsection {section}{1}{\z@}%
                                   {-3.5ex \@plus -1ex \@minus -.2ex}%
                                   {2.3ex \@plus.2ex}%
                                   {\normalfont\bfseries}}

\appendix
\begin{center}
	\textbf{\Large Appendix}
\end{center}

Additional results and discussions of our supplementary are as follows:
\begin{itemize}
    \item Section \ref{sec:suple_a}: Additional details about Foggy KITTI dataset and MonoWAD.
    \item Section \ref{sec:suple_b}: Additional results on KITTI 3D dataset (\textit{i.e.,} weather-robustness, BEV results, fog density).
    \item Section \ref{sec:suple_c}: Additional results on Virtual KITTI dataset.
    \item Section \ref{sec:suple_d}: Additional results on Real-World dataset.
    \item Section \ref{sec:suple_e}: Qualitative comparison with dehazing method.
    \item Section \ref{sec:suple_f}: Additional visualization results.
    \item Section \ref{sec:suple_g}: Video demo.
\end{itemize}

\vspace{-0.5cm}

%------------------------------------ Figure 1
%###############################################################################################
\begin{figure*}[h]
    \begin{minipage}[b]{1.0\linewidth}
    \centering
    \centerline{\includegraphics[width=12.2cm]{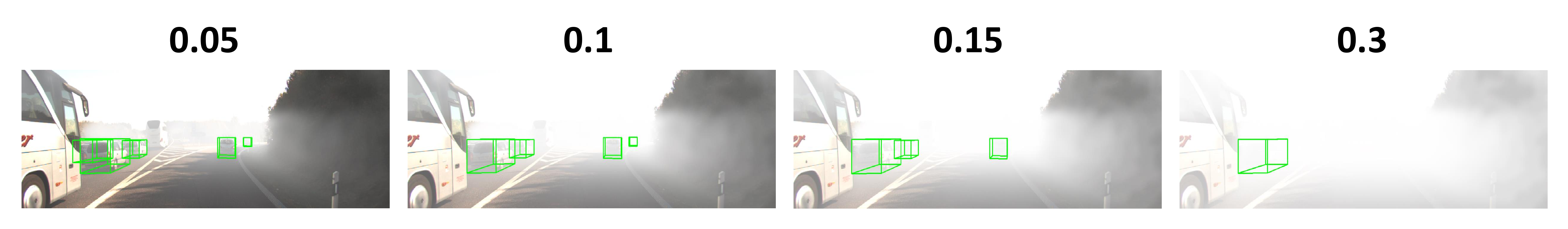}}
    \end{minipage}
    \vspace{-0.7cm}
    \caption{Examples with fog densities $\delta=\{0.05, 0.1, 0.15, 0.3\}$.}
    \label{fig:sup_1}
    \vspace{-0.8cm}
\end{figure*}
%###############################################################################################

\section{Additional Details about Foggy KITTI Dataset and MonoWAD}\label{sec:suple_a}
\subsubsection{Foggy KITTI Dataset.}
Given image $I$, we adopt pre-trained DORN \cite{DORN} to obtain depth map $I_D$ and calculate transmittance $T(I_D,\delta)$ using $I_D$ and fog density $\delta$. After estimating atmospheric light $I_A$ from $I$, foggy KITTI is obtained via Eq. \ref{supl_eq:1}. Following \cite{SeeingThroughFog, Foggy_Cityscapes}, we can generate various foggy images via $\delta$=$\{${0.05,0.1,0.15, 0.3}$\}$ (Fig. \ref{fig:sup_1}). In all experiments of our main paper, we set a fog density $\delta$=0.1.

\begin{equation}
I_F = (I*T(I_D, \delta) + I_A*(1-T(I_D, \delta)).
\label{supl_eq:1}
\end{equation}
In addition, unlike the Multifog KITTI dataset \cite{Fog_3dod}, our foggy KITTI utilizes depth information inferred from monocular images to generate photo-realistic fog data for monocular 3D object detection. Moreover, the various densities are provided separately, rather than integrated. %offering greater flexibility in research.

\subsubsection{MonoWAD in Clear Weather.}
In the training process, our weather codebook (WC) and weather-adaptive diffusion model (WAD) learn clear features via clear knowledge recalling (CKR) loss $\mathcal{L}_{ckr}$ and weather-adaptive enhancement loss $\mathcal{L}_{wae}$ to enhance the weather and emphasize feature by cross attention, and detection loss $\mathcal{L}_{OD}$ to enhance the features of the backbone for detection. This is different from performing detection by dehazing fog, as it serves to remove fog while emphasizing features. It also dynamically enhances the feature representation of input images (clear or foggy), allowing it to perform robustly in both clear and foggy weather conditions.

\subsubsection{Details of Weather Codebook.}
We employ a single weather codebook in our MonoWAD. The weather codebook has 1.05M parameters, which is 1.9\% of the total 54.25M parameters in the baseline model. With a single codebook, ours can learn to memorize the knowledge of clear weather using the clear knowledge recalling (CKR) loss $\mathcal{L}_{ckr}$ and generate reference features for other weather conditions (\textit{e.g.,} foggy, rainy, snowy) (Eq. \ref{eq:3} of our main paper).

%------------------------------------ Figure 2
%###############################################################################################
\begin{figure*}[t]
    \begin{minipage}[b]{1.0\linewidth}
    \centering
    \centerline{\includegraphics[width=12.2cm]{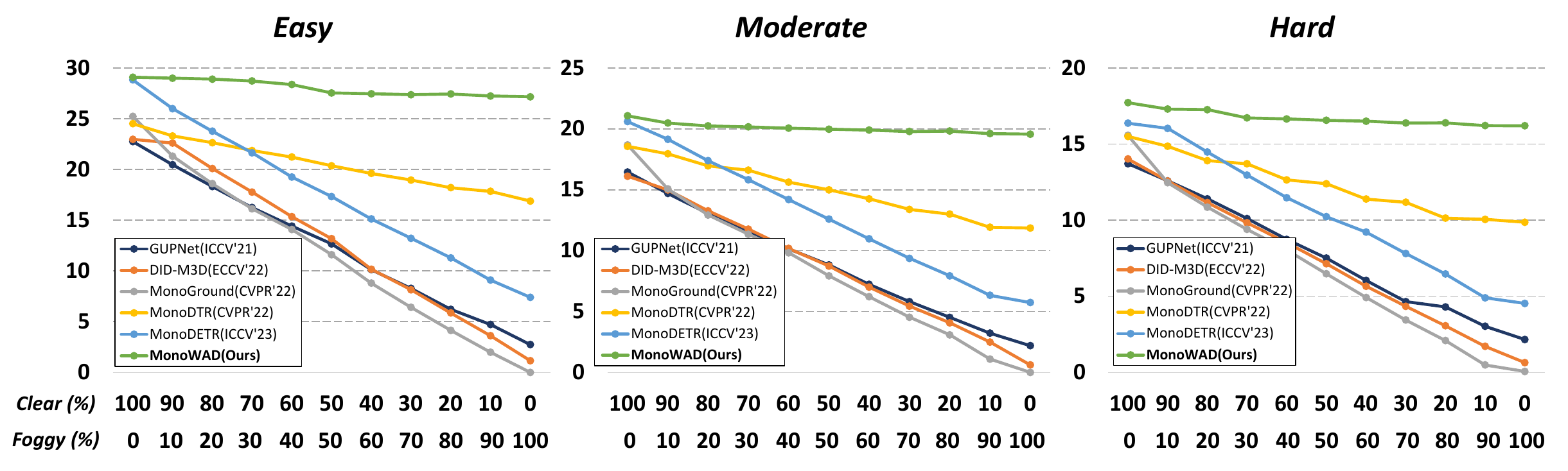}}
    \end{minipage}
    \vspace{-0.7cm}
    \caption{Performance variations of car category on KITTI validation set under various weather conditions, including foggy weather (foggy) and clear weather (clear) based on its percentage. `Clear (n\%) and Foggy (m\%)' indicates that n\% images of the validation set correspond to clear weather, and m\% images correspond to foggy weather.}
    \label{fig:per_supp}
    \vspace{-0.5cm}
\end{figure*}
%###############################################################################################

\subsubsection{Details of Weather-Adaptive Diffusion Model.}
We further provide a more detailed explanation of our weather-adaptive diffusion model, including the noise in the forward process and enhancement in the reverse process. In the forward process, fog distribution $\mathcal{F}=x^{f}-x^{c}$ is the difference between clear and fog features, used as our diffusion noise. Fog variant $\epsilon_{n}$ is applied based on a fixed Markov Chain of $T$ timesteps determined by variance schedule $\beta_t$. During inference, our diffusion model estimates the mean $\boldsymbol{\mu}_\theta$ and variance $\boldsymbol{\Sigma}_\theta$ at each timestep, and $\mathcal{F}$ is estimated by aggregating them across all timesteps. Following \cite{DDPM}, we set variance $\boldsymbol{\Sigma}_\theta(x^c_t, t)$ to be $\sigma_t^2 \boldsymbol{I}$, where $\sigma_t^2=\beta_t$. In the reverse process, the weather-adaptive diffusion model consists of an autoencoder (U-Net) that has an encoder/mid-block/decoder with no additional backbone. In this architecture, cross-attention between the mid-block feature and the weather-reference feature from weather codebook is conducted. It takes the previous step $x^c_T$ as input and predicts the next step $x^c_{T-1}$. As shown in Fig. \ref{fig:diffusion} of our main paper, we operate the same autoencoder at different timesteps to gradually enhance the representation from $x^c_T$ to $x^c_0$. We trained the weather codebook, weather-adaptive diffusion model, and detection block as a single model in an end-to-end manner, without requiring any additional data.

\section{Additional Results on KITTI 3D Dataset}\label{sec:suple_b}
\subsubsection{Weather-Robustness Experiments.}
As we have mentioned mixed foggy and clear weather conditions as an extension of the weather-robustness experiments in Section 4.3 (Results on KITTI 3D Dataset) of our main paper, we further compared the 3D detection performance under the clear and foggy validation set based on its percentage (Clear/Foggy: 100\%/0\% to 0\%/100\% balancing in 10\% intervals). We conduct experiments by selecting random images from both foggy and clear weather according to predetermined seeds, ensuring that all models are tested under identical conditions. The results are shown in Fig. \ref{fig:per_supp}. As the ratio of the foggy increased, the performance of the existing methods gradually decreased. For example, when the clear/foggy ratio changed from 70\%/30\% (Table \ref{table:percentage3}) to 30\%/70\% (Table \ref{table:percentage3}), the performance of MonoDETR dropped significantly from (21.65, 15.83, and 12.97) to (13.22, 9.37, and 7.81) for (`Easy', `Moderate', and `Hard') settings, respectively. In contrast, the performance change of our method is marginal even when we vary the ratios of clear and foggy conditions. The experimental results demonstrate the weather-robustness property of our method.

%------------------------------------ Table 1
%##################################################################################################
\begin{table}[tb]
    \caption{Performance (AP$_{3D}$) variations of car category on KITTI validation set under weather conditions, including foggy weather (foggy) and clear weather (clear) based on its percentage. `Clear(n\%)+Foggy(m\%)' indicates that n\% images of the validation set correspond to clear weather, and m\% images correspond to foggy weather.}
    \vspace{-0.6cm}
    \renewcommand{\tabcolsep}{3.5mm}
	\centering
	\begin{center} 
		\resizebox{0.999\linewidth}{!}{
		  \begin{tabular}{c ccc ccc ccc}
				\Xhline{3\arrayrulewidth}
				\rule{0pt}{10.5pt} \multirow{3}{*}[-0.4em]{\bf Method} & \multicolumn{3}{c}{\bf Clear(70\%)+Foggy(30\%)} & \multicolumn{3}{c}{\bf Clear(50\%)+Foggy(50\%)} & \multicolumn{3}{c}{\bf Clear(30\%)+Foggy(70\%)} \\\cmidrule(lr){2-4}\cmidrule(lr){5-7}\cmidrule(lr){8-10}
				& \bf Easy & \bf Mod. & \bf Hard & \bf Easy & \bf Mod. & \bf Hard & \bf Easy & \bf Mod. & \bf Hard \\\hline
				\rule{0pt}{10.5pt}
				GUPNet \cite{GUPNet} (ICCV'21) 
				& 16.25 & 11.59 & 10.11 & 12.67 & 8.83 & 7.51 & 8.28 & 5.81 & 4.65 \\
				DID-M3D \cite{DID-M3D} (ECCV'22) 
				& 17.78 & 11.77 & 9.83 & 13.18 & 8.74 & 7.14 & 8.14 & 5.44 & 4.34 \\
				MonoGround \cite{MonoGround} (CVPR'22)
				& 16.13 & 11.35 & 9.40 & 11.61 & 7.93 & 6.48 & 6.41 & 4.53 & 3.44 \\
				MonoDTR\cite{MonoDTR} (CVPR'22) 
				& 21.87 & 16.61 & 13.71 & 20.37 & 15.00 & 12.40 & 18.96 & 13.39 & 11.18 \\
				MonoDETR \cite{MonoDETR} (ICCV'23) 
				& 21.65 & 15.83 & 12.97 & 17.33 & 12.59 & 10.23 & 13.22 & 9.37 & 7.81 \\
				\cdashline{1-10}
				\rule{0pt}{10.8pt} \bf MonoWAD (Ours) 
				& \bf 28.73 & \bf 20.17 & \bf 16.73 & \bf 27.55 & \bf 19.98 & \bf 16.57 & \bf 27.38 & \bf 19.79 & \bf 16.39 \\
				\Xhline{3\arrayrulewidth}
		  \end{tabular}
		}
        \vspace{-0.4cm}
		\label{table:percentage3}
	\end{center}
\end{table}
%##################################################################################################

%------------------------------------ Table 2
%##################################################################################################
\begin{table}[t]
    \caption{Detection results (AP$_{BEV}$) of car category on KITTI validation set under foggy weather and clear weather conditions. \textbf{Bold}/\underline{underlined} fonts indicate the best/second-best results.}
    \vspace{-0.6cm}
    \renewcommand{\tabcolsep}{3.5mm}
	\centering
	\begin{center} 
		\resizebox{0.999\linewidth}{!}
		{
			\begin{tabular}{c ccc ccc ccc}
				\Xhline{3\arrayrulewidth}
                \rule{0pt}{10.5pt} \multirow{2}{*}[-0.2em]{\bf Method} & \multicolumn{3}{c}{\bf Foggy (\bf AP$_{BEV}$)} & \multicolumn{3}{c}{\bf Clear (\bf AP$_{BEV}$)} & \multicolumn{3}{c}{\multirow{1}{*}{\bf Average}} \\\cmidrule(lr){2-4}\cmidrule(lr){5-7}\cmidrule(lr){8-10}
				& \bf Easy & \bf Mod. & \bf Hard & \bf Easy & \bf Mod. & \bf Hard & \bf Easy & \bf Mod. & \bf Hard\\\hline
				\rule{0pt}{10.5pt}
				GUPNet \cite{GUPNet} (ICCV'21) 
				& 5.13 & 4.37 & 2.93 & 31.07 & 22.94 & 19.75 & 18.10 & 13.66  & 11.34 \\
				DID-M3D \cite{DID-M3D} (ECCV'22) 
				& 2.40 & 1.78 & 0.86 & 31.10 & 22.76 & 19.50 & 16.75 & 12.27 & 10.18\\
				MonoGround \cite{MonoGround} (CVPR'22) 
				& 0.00 & 0.00 & 0.07 & 32.68 & 24.79 & 20.56 & 16.34 & 6.20 & 10.32\\
				MonoDTR\cite{MonoDTR} (CVPR'22) 
				& \underline{22.01} & \underline{14.84} & \underline{12.74} & 33.33 & 25.35 & 21.68 & \underline{27.67} & \underline{20.10} & \underline{17.21} \\
				MonoDETR \cite{MonoDETR} (ICCV'23) 
				& 11.03 & 7.26 & 5.69 & \underline{37.86} & \underline{26.95} & \underline{22.80} & 18.12 & 17.11 & 14.25 \\
				\cdashline{1-10}
				\rule{0pt}{10.8pt} \bf MonoWAD (Ours)
				& \bf 35.70 & \bf 25.31 & \bf 21.43 & \bf 38.07 & \bf 26.97 & \bf 23.04 & \bf 36.89 & \bf 26.14 & \bf 22.24\\
				\Xhline{3\arrayrulewidth}
			\end{tabular}
		}
        \vspace{-0.4cm}
	\label{table:valid_bev}
	\end{center}
\end{table}
%##################################################################################################

%------------------------------------ Table 3
%##################################################################################################
\begin{table}[t]
    \caption{Detection results (AP$_{3D}$) of car category on foggy KITTI validation set under various foggy densities $\delta=\{0.05, 0.15, 0.3\}$ ($\delta=0.1$ is in main paper). The results of the state-of-the-art methods under foggy weather are obtained through our reproduction with the official source code. \textbf{Bold}/\underline{underlined} fonts indicate the best/second-best results.}
    \vspace{-0.6cm}
    \renewcommand{\tabcolsep}{3.5mm}
	\centering
	\begin{center} 
		\resizebox{0.999\linewidth}{!}
		{
			\begin{tabular}{c ccc ccc ccc}
				\Xhline{3\arrayrulewidth}
                \rule{0pt}{10.5pt} \multirow{2}{*}[-0.2em]{\bf Method} & \multicolumn{3}{c}{\bf $\delta=0.05$} & \multicolumn{3}{c}{\bf $\delta=0.15$} & \multicolumn{3}{c}{\multirow{1}{*}{\bf $\delta=0.3$}} \\\cmidrule(lr){2-4}\cmidrule(lr){5-7}\cmidrule(lr){8-10}
				& \bf Easy & \bf Mod. & \bf Hard & \bf Easy & \bf Mod. & \bf Hard & \bf Easy & \bf Mod. & \bf Hard\\\hline
				\rule{0pt}{10.5pt}
				GUPNet \cite{GUPNet} (ICCV'21) 
				& 7.29 & 5.16 & 4.16 & 0.64 & 0.93 & 0.88 & 0.00 & 0.00 & 0.00\\
				DID-M3D \cite{DID-M3D} (ECCV'22) 
				& 9.66 & 6.90 & 5.46 & 0.50 & 0.74 & 0.77 & 0.00 & 0.00 & 0.00\\
				MonoGround \cite{MonoGround} (CVPR'22) 
				& 0.53 & 0.28 & 0.31 & 0.00 & 0.00 & 0.00 & 0.00 & 0.00 & 0.00\\
				MonoDTR\cite{MonoDTR} (CVPR'22) 
				& \underline{22.42} & \underline{16.24} & \underline{13.09} & \underline{11.38} & \underline{7.27} & \underline{5.74} & \underline{2.24} & \underline{1.89} & \underline{1.85}\\
				MonoDETR \cite{MonoDETR} (ICCV'23) 
				& 15.06 & 10.70 & 8.89 & 3.61 & 2.92 & 2.02 & 0.36 & 0.36 & 0.36\\
				\cdashline{1-10}
				\rule{0pt}{10.8pt} \bf MonoWAD (Ours)
				& \bf 26.99 & \bf 19.19 & \bf 15.88 & \bf 15.48 & \bf 10.71 & \bf 8.60 & \bf 9.66 & \bf 6.90 & \bf 5.46\\
				\Xhline{3\arrayrulewidth}
			\end{tabular}
		}
        \vspace{-0.4cm}
	\label{table:valid_density}
	\end{center}
\end{table}
%##################################################################################################

\subsubsection{BEV Results on KITTI validation set.}
We further compared the AP$_{BEV}$ on KITTI \cite{KITTI} and foggy KITTI validation set in Table \ref{table:valid_bev}. Similar to Table \ref{table:valid} of our main paper, our MonoWAD outperforms the existing monocular 3D object detector under clear and foggy weather.

\subsubsection{Results under Different Fog Density.} We further compared the AP$_{3D}$ on foggy KITTI validation set under different fog density $\delta=\{0.05,0.15,0.3\}$. As shown in Table \ref{table:valid_density}, even as the fog density $\delta$ increases, ours still outperforms the state-of-the-art methods. In Fig. \ref{fig:sup_1}, we also visualize the 3D detection results on foggy KITTI images of various fog densities, demonstrating the robustness of our method under different visibility conditions.

%------------------------------------ Table 4
%##################################################################################################
\begin{table}[t]
    \caption{Detection results (AP$_{3D}$) of car category on Virtual KITTI under foggy, rainy, and sunset conditions, are based on an equal percentage mix of these weather conditions. \textbf{Bold}/\underline{underlined} fonts indicate the best/second-best results.}
    \vspace{-0.6cm}
    \renewcommand{\tabcolsep}{4.2mm}
	\centering
	\begin{center}
		\resizebox{0.85\linewidth}{!}
		{
			\begin{tabular}{c ccc ccc}
				\Xhline{3\arrayrulewidth}
				\rule{0pt}{10pt} \multirow{3}{*}[-0.1em]{\bf Method} & \multicolumn{3}{c}{\bf Foggy/Rainy/Sunset (33.3\%)} & \multicolumn{3}{c}{\bf Foggy/Rainy/Sunset (33.3\%)} \\\cmidrule(lr){2-4}\cmidrule(lr){5-7}
				& \bf Easy & \bf Mod. & \bf Hard & \bf Easy & \bf Mod. & \bf Hard \\\hline
				\rule{0pt}{9.5pt}
				GUPNet \cite{GUPNet} (ICCV'21)
				& 2.29 & 1.21 & 1.19 & 9.76 & 5.58 & 5.56\\
				DID-M3D \cite{DID-M3D} (ECCV'22)
				& 0.40 & 0.13 & 0.13 & 5.37 & 3.25 & 3.21\\
				MonoGround \cite{MonoGround} (CVPR'22)
				& 4.39 & 2.50 & 2.43 & 17.27 & 11.29 & 11.21\\
				MonoDTR \cite{MonoDTR} (CVPR'22)
				& \underline{10.27} & \underline{5.88} & \underline{5.84} & \underline{22.09} & \underline{14.24} & \underline{14.21}\\
				MonoDETR \cite{MonoDETR} (ICCV'23)
				& 6.17 & 3.31 & 3.28 & 15.84 & 9.77 & 9.79\\
				\cdashline{1-7}
				\rule{0pt}{10.8pt} \bf MonoWAD (Ours) 
				& \bf 13.69 & \bf 8.22 & \bf 8.14 & \bf 29.46 & \bf 18.81 & \bf 18.76\\
				\Xhline{3\arrayrulewidth}
			\end{tabular}
		}
        \vspace{-0.7cm}
		\label{table:vir_robust}
	\end{center}
\end{table}
%##################################################################################################

%------------------------------------ Table 5
%##################################################################################################
\begin{table}[t]
    \caption{Detection results (AP$_{3D}$) of car category on Seeing Through Fog under various weather conditions (\textit{e.g.,} clear, foggy, rainy, snowy). \textbf{Bold}/\underline{underlined} fonts indicate the best/second-best results.}
    \vspace{-0.6cm}
    \renewcommand{\tabcolsep}{3.5mm}
	\centering
	\begin{center} 
		\resizebox{0.999\linewidth}{!}
		{
			\begin{tabular}{c ccc ccc ccc ccc}
				\Xhline{3\arrayrulewidth}
				\rule{0pt}{10pt} \multirow{2}{*}[-0.1em]{\bf Method}  & \multicolumn{3}{c}{\bf Clear (AP$_{3D}$)} & \multicolumn{3}{c}{\bf Foggy (AP$_{3D}$)} & \multicolumn{3}{c}{\bf Rainy (AP$_{3D}$)} & \multicolumn{3}{c}{\bf Snowy (AP$_{3D}$)}
				\\\cmidrule(lr){2-4}\cmidrule(lr){5-7}\cmidrule(lr){8-10}\cmidrule(lr){11-13}
				& \bf Easy & \bf Mod. & \bf Hard & \bf Easy & \bf Mod. & \bf Hard & \bf Easy & \bf Mod. & \bf Hard & \bf Easy & \bf Mod. & \bf Hard \\\hline
				\rule{0pt}{9.5pt}
				MonoDTR \cite{MonoDTR}
				& 10.08 & 8.71 & 6.98 & 19.26 & 16.66 & 15.37 & 5.30 & 4.99 & 3.53 & 9.05 & 7.24 & 6.35\\
                MonoDTR + RIDCP \cite{RIDCP}
				& 9.44 & 8.57 & 6.95 & 17.22 & 14.48 & 13.48 & 3.85 & 4.32 & 3.67 & 8.12 & 6.66 & 5.28\\
                MonoDTR + ZeroScatter \cite{zeroscatter}
				& 7.54 & 7.08 & 5.68 & 13.30 & 11.99 & 10.89 & 3.27 & 3.47 & 2.77 & 6.15 & 5.25 & 4.62\\ \hline
				MonoDETR \cite{MonoDETR}
				& \underline{17.09} & \underline{12.26} & \underline{9.49} & \underline{26.78} & \underline{18.44} & \underline{16.41} & \underline{11.12} & \underline{7.09} & \underline{5.39} & \underline{15.94} & \underline{10.20} & \underline{8.66}\\
                MonoDETR + RIDCP \cite{RIDCP}
				& 16.66 & 11.07 & 9.19 & 25.05 & 17.52 & 15.67 & 9.83 & 6.24 & 4.96 & 14.92 & 9.69 & 8.18\\
                MonoDETR + ZeroScatter \cite{zeroscatter}
				& 14.05 & 10.22 & 7.61 & 19.47 & 13.61 & 12.07 & 6.39 & 4.16 & 3.14 & 11.70 & 7.87 & 6.54\\
				\cdashline{1-13}
				\rule{0pt}{10.8pt} \bf MonoWAD (Ours) 
				& \bf 20.44 & \bf 14.24 & \bf 10.95 & \bf 30.31 & \bf 20.51 & \bf 18.68 & \bf 15.10 & \bf 9.15 & \bf 6.86 & \bf 19.04 & \bf 12.03 & \bf 10.19\\
				\Xhline{3\arrayrulewidth}
			\end{tabular}
		}
        \vspace{-0.4cm}
	\label{table:supp_real}
	\end{center}
\end{table}
%##################################################################################################

\section{Additional Results on Virtual KITTI Dataset}\label{sec:suple_c}
\subsubsection{Weather-Robustness Experiments.}
We also conducted a weather-robustness experiment under mixed foggy, rainy, and sunset weather conditions. Same as Table \ref{table:percentage3}, we select random images, and we compared 3D detection performance under mixed weather conditions based on an equal percentage (percentage: 33.3\%). As shown in Table \ref{table:vir_robust}, our MonoWAD outperforms the existing method in the coexisting of various weather conditions. These results demonstrate that our MonoWAD is still robust and insensitive to various weather conditions that can be faced in real-world autonomous driving.

\section{Additional Results on Real-World Dataset}\label{sec:suple_d}
We investigate the transferability to real-world conditions of our method compared to the application of other enhancement methods \cite{RIDCP, zeroscatter} on two state-of-the-art detectors \cite{MonoDTR, MonoDETR}. To this end, we compared the AP$_{3D}$ on real-world images from the Seeing Through Fog dataset \cite{SeeingThroughFog}. As shown in Table \ref{table:supp_real}, our MonoWAD consistently outperforms them in various weather, demonstrating its transferability to real-world conditions.

%------------------------------------ Figure 3
%###############################################################################################
\begin{figure*}[t]
    \begin{minipage}[b]{1.0\linewidth}
    \centering
    \centerline{\includegraphics[width=12.2cm]{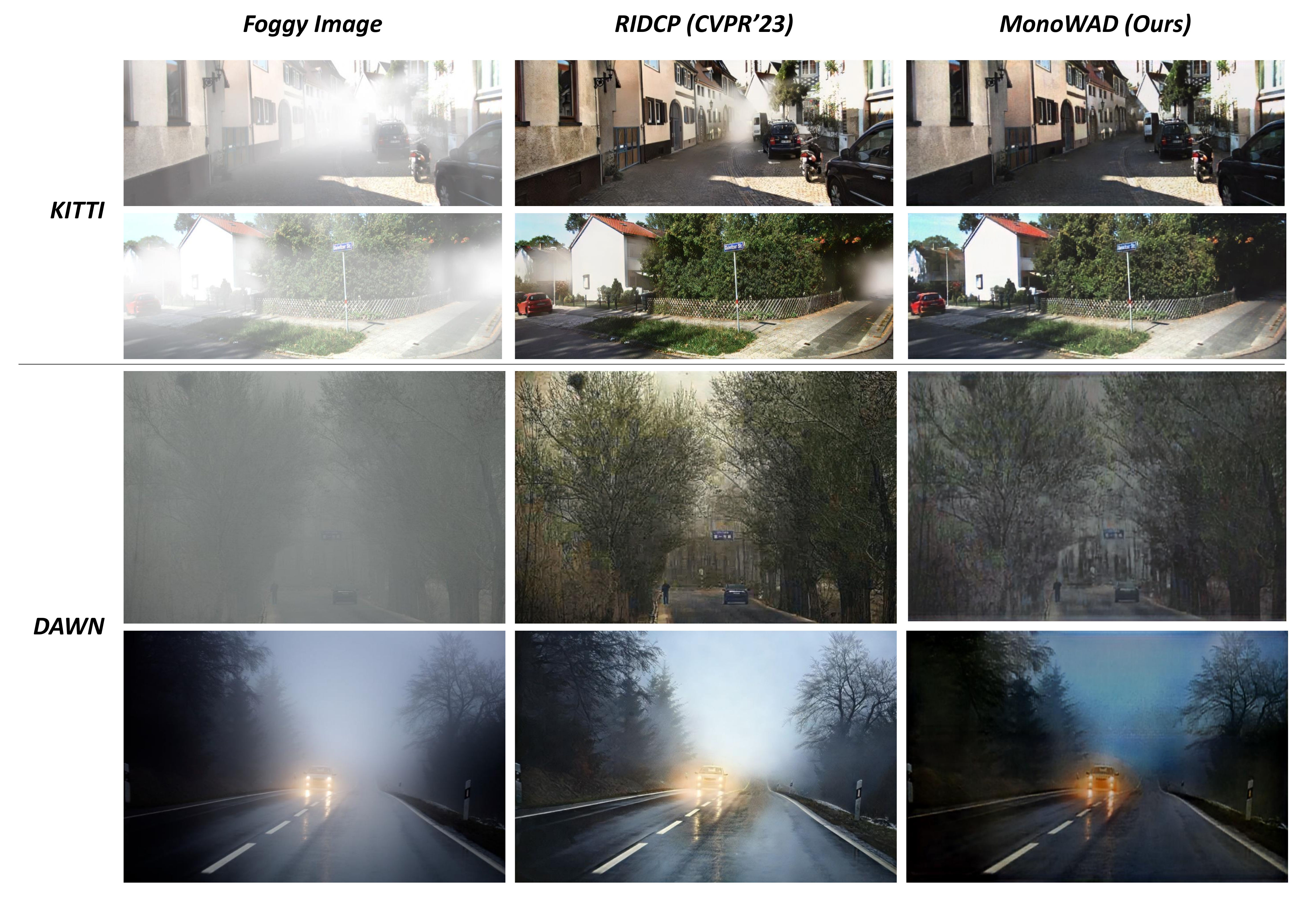}}
    \end{minipage}
    \vspace{-0.7cm}
    \caption{Qualitative comparison on KITTI and DAWN dataset.}
    \label{fig:sup_fig4}
    \vspace{-0.15cm}
\end{figure*}
%###############################################################################################

%------------------------------------ Figure 4
%###############################################################################################
\begin{figure*}[t]
    \begin{minipage}[b]{1.0\linewidth}
    \centering
    \centerline{\includegraphics[width=12.2cm]{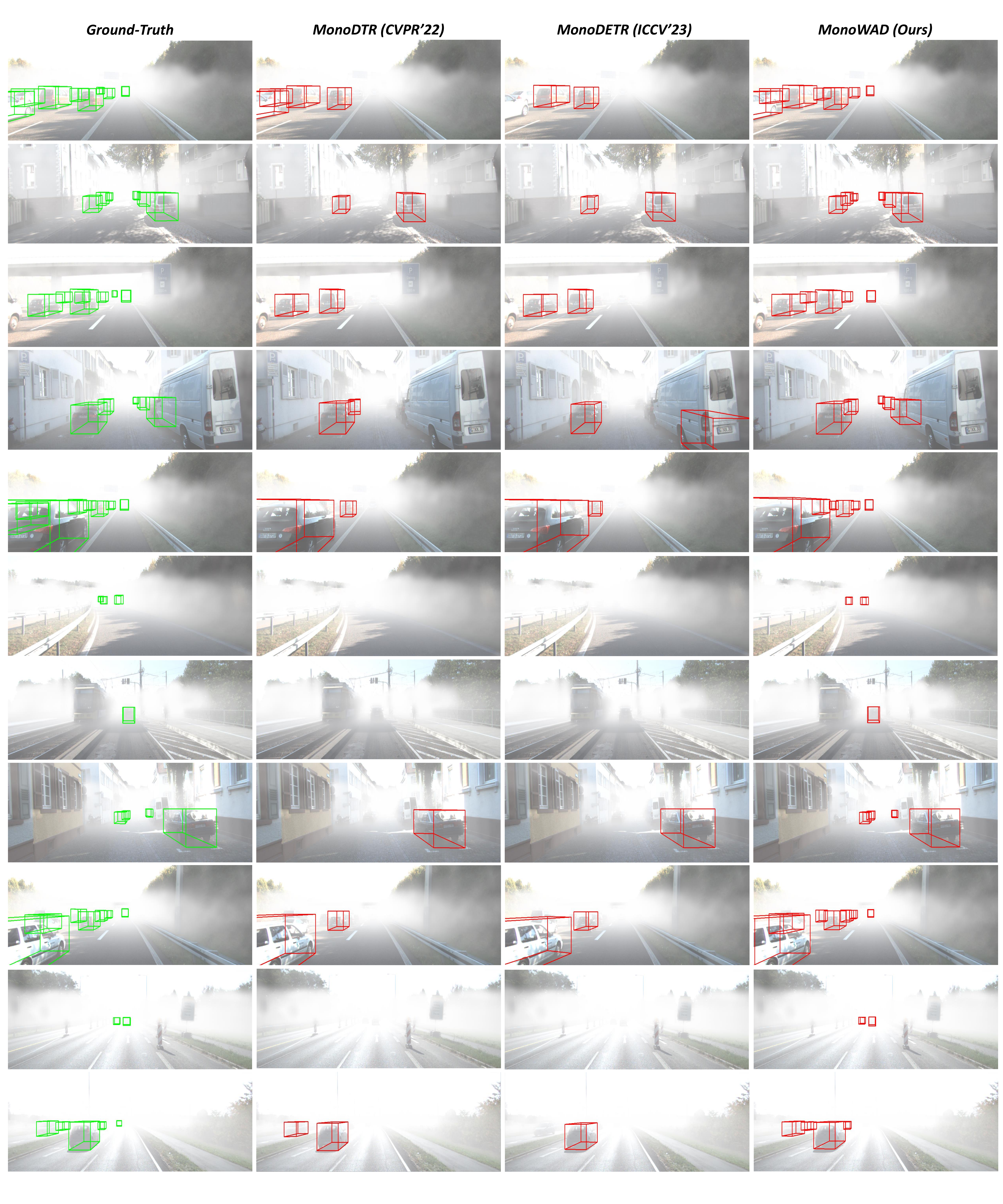}}
    \end{minipage}
    \vspace{-0.7cm}
    \caption{Comparison of 3D detection examples on foggy KITTI dataset (green: ground-truth, red: predicted 3D bounding-box) between our MonoWAD and two detectors, MonoDTR \cite{MonoDTR} and MonoDETR \cite{MonoDETR}, that show the most improved performances among existing methods.}
    \label{fig:sup_fig2}
    \vspace{-0.15cm}
\end{figure*}
%###############################################################################################

%------------------------------------ Figure 5
%###############################################################################################
\begin{figure*}[t]
    \begin{minipage}[b]{1.0\linewidth}
    \centering
    \centerline{\includegraphics[width=12.2cm]{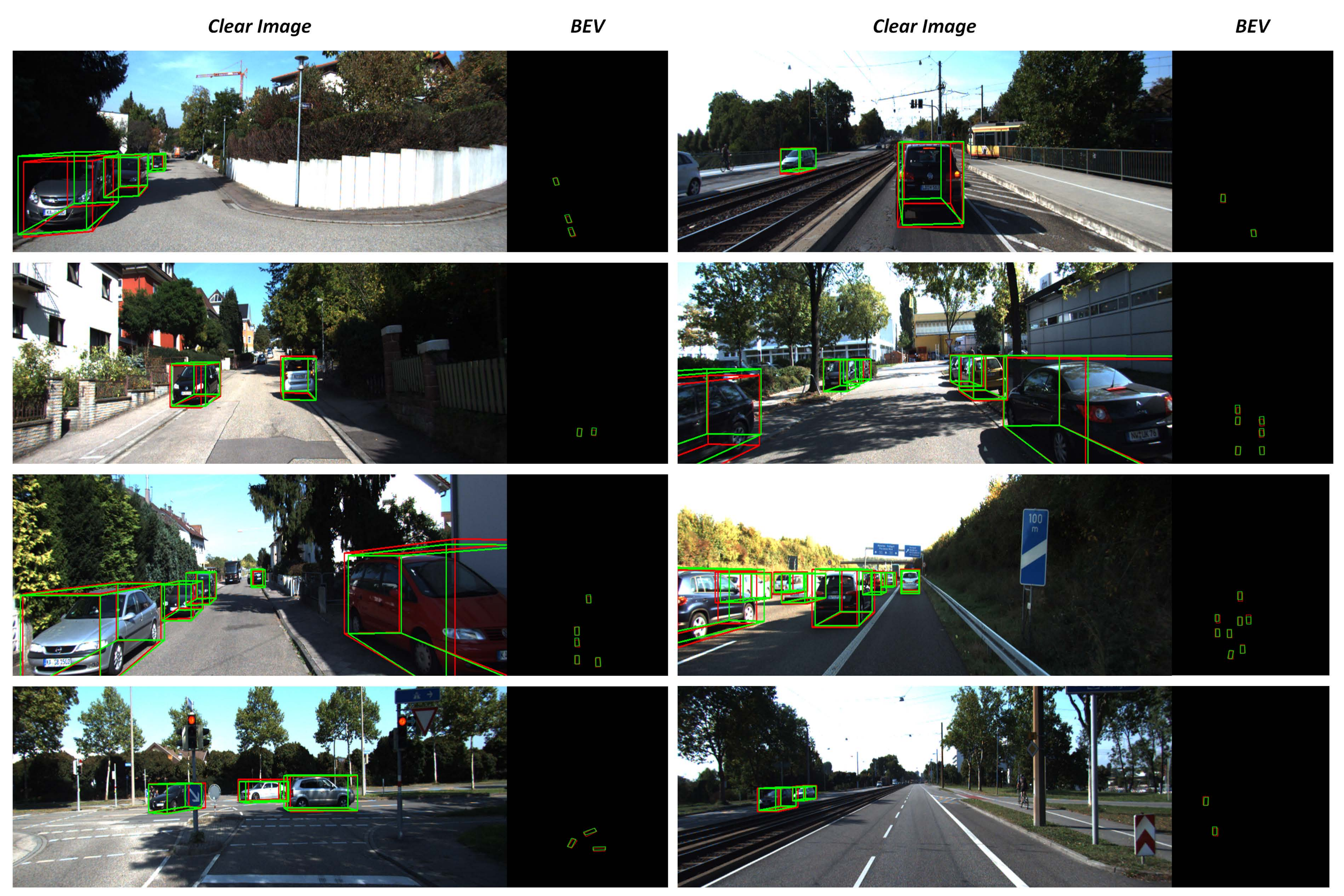}}
    \end{minipage}
    \vspace{-0.7cm}
    \caption{Comparison of 3D detection examples in the image plane and BEV plane under clear weather KITTI dataset (red: ground-truth, green: predicted 3D bounding-box of our MonoWAD).}
    \label{fig:sup_fig3}
    \vspace{-0.15cm}
\end{figure*}
%###############################################################################################

%------------------------------------ Figure 6
%###############################################################################################
\begin{figure*}[t]
    \begin{minipage}[b]{1.0\linewidth}
    \centering
    \centerline{\includegraphics[width=12.2cm]{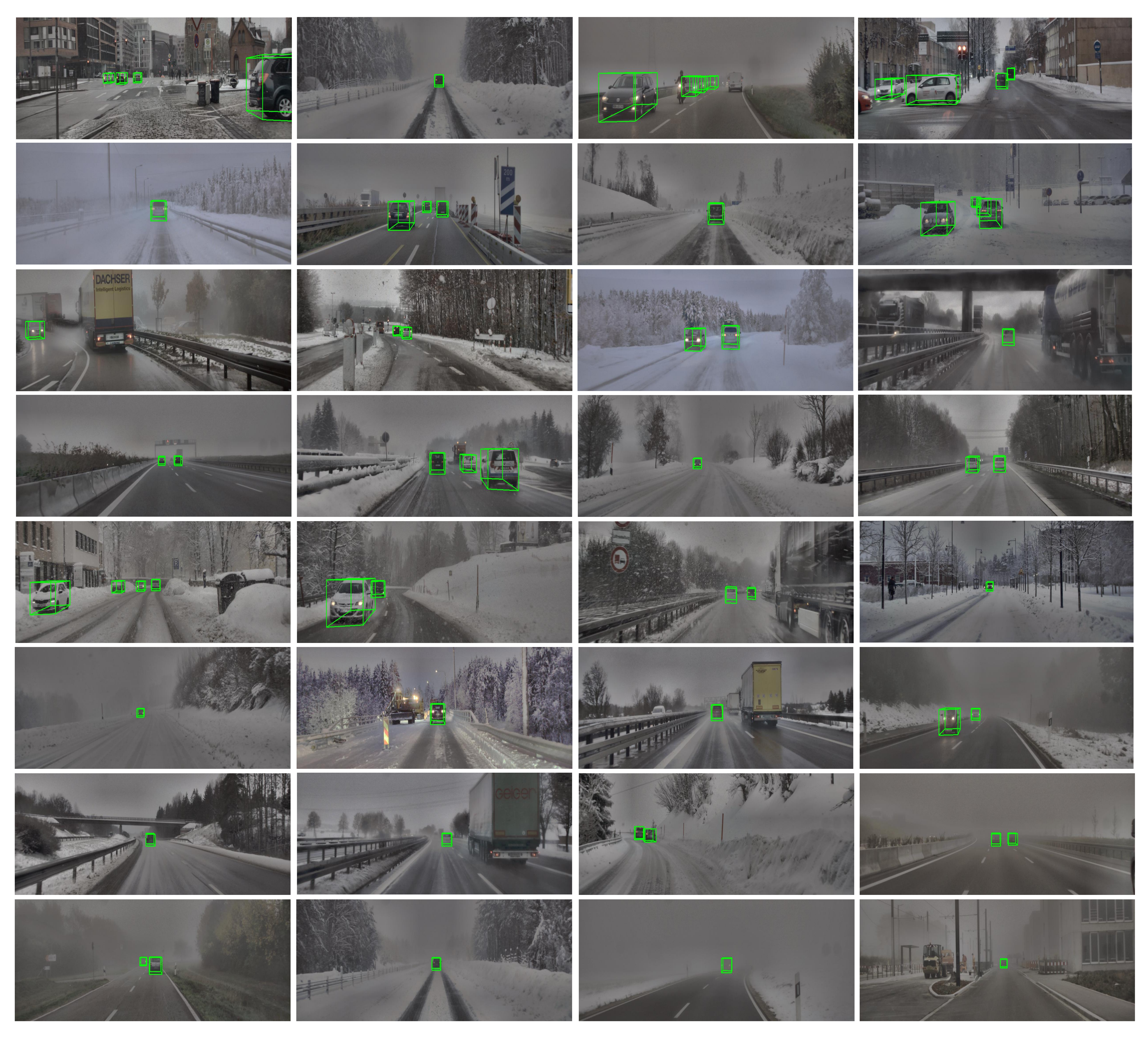}}
    \end{minipage}
    \vspace{-0.7cm}
    \caption{3D detection results on real-world images of various weather conditions (\textit{e.g.,} foggy, rainy, snowy).}
    \label{fig:sup_fig5}
    \vspace{-0.15cm}
\end{figure*}
%###############################################################################################

\section{Qualitative Comparison with Dehazing Method}\label{sec:suple_e}
In Section 4.5 (Comparison with Dehazing Methods) of our main paper, we applied dehazing method to an existing monocular 3D object detector \cite{MonoDTR}. Therefore, we show the results of the state-of-the-art image dehazing method, RIDCP \cite{RIDCP}, to the foggy KITTI validation set in Fig. \ref{fig:sup_fig4}. We further show the results of our MonoWAD in the dehazing application. Since our MonoWAD is designed for a weather-robust monocular 3D object detector, we performed dehaze by adding a simple decoder architecture to our weather-adaptive diffusion and weather codebook. Fig. \ref{fig:sup_fig4} further demonstrates that MonoWAD is effective not only on the foggy KITTI dataset but also on the DAWN dataset \cite{dawn}, which contains real foggy image from real-world scenarios. This shows that our proposed method for dynamically enhancing the feature representation of the input images according to the weather conditions works well and has the potential to be applied to other tasks beyond monocular 3D object detection.

\section{Additional Visualization Results}\label{sec:suple_f}
\noindent{\textbf{Foggy Weather.}} We further show the 3D detection results in foggy weather to compare our MonoWAD with MonoDTR \cite{MonoDTR} and MonoDETR \cite{MonoDETR}, which exhibit the highest performance among existing methods \cite{GUPNet, DID-M3D, MonoGround, MonoDTR, MonoDETR} under various foggy scenarios. The results are shown in Fig. \ref{fig:sup_fig2}. The results demonstrate that the proposed MonoWAD effectively detects objects obscured by fog compared to existing methods. \\

\noindent{\textbf{Clear Weather.}}  We also visualize the 3D detection results in clear weather to compare our MonoWAD (green) with ground-truth annotations (red).  As shown in Fig. \ref{fig:sup_fig3}, the proposed MonoWAD effectively detects objects even in various scenes under clear weather conditions. \\

\noindent{\textbf{Diverse Weathers on Real-World Images.}}  We also visualize the 3D detection results in diverse weather conditions (\textit{i.e.,} foggy, rainy, snowy) using real-world images from Seeing Through Fog dataset \cite{SeeingThroughFog}.  As shown in Fig. \ref{fig:sup_fig5}, the proposed MonoWAD effectively detects objects even in various scenes under diverse weather conditions. \\

\section{Video Demo}\label{sec:suple_g}
We provide video materials to show the detection results of our method and existing methods under various weather conditions (clear and foggy). Please see the video in \href{https://github.com/VisualAIKHU/MonoWAD}{our official repository}.

\end{document}